\documentclass[letterpaper]{article} % DO NOT CHANGE THIS
\usepackage{aaai2026}  % DO NOT CHANGE THIS
\nocopyright
\usepackage{times}  % DO NOT CHANGE THIS
\usepackage{helvet}  % DO NOT CHANGE THIS
\usepackage{courier}  % DO NOT CHANGE THIS
\usepackage[hyphens]{url}
\usepackage{graphicx} % DO NOT CHANGE THIS
\def\UrlFont{\rm}  % DO NOT CHANGE THIS
\usepackage{natbib}  % DO NOT CHANGE THIS AND DO NOT ADD ANY OPTIONS TO IT
\usepackage{caption} % DO NOT CHANGE THIS AND DO NOT ADD ANY OPTIONS TO IT
\usepackage{comment}
\usepackage{algorithm}
\usepackage{algorithmic}
\usepackage{amsmath}
\usepackage{newfloat}
\usepackage{listings}
\usepackage{booktabs}
\DeclareCaptionStyle{ruled}{labelfont=normalfont,labelsep=colon,strut=off} % DO NOT CHANGE THIS
\floatstyle{ruled}
\newfloat{listing}{tb}{lst}{}
\floatname{listing}{Listing}
\PassOptionsToPackage{table}{xcolor}
\usepackage{colortbl} 
\title{Causally Steered Diffusion for Automated Video
Counterfactual Generation}

\author {
    Nikos Spyrou\textsuperscript{\rm 1,\rm 2,\rm 3},
    Athanasios Vlontzos\textsuperscript{\rm 7}\thanks{Work conducted while at Spotify, UK},
    Paraskevas Pegios\textsuperscript{\rm 4,\rm 5},
    Thomas Melistas\textsuperscript{\rm 1,\rm 2,\rm 3},\\
    Nefeli Gkouti\textsuperscript{\rm 1,\rm 2,\rm 3},
    Yannis Panagakis\textsuperscript{\rm 1,\rm 2},
    Giorgos Papanastasiou\textsuperscript{\rm 2,\rm 6},
    Sotirios A. Tsaftaris\textsuperscript{\rm 2,\rm 3}
}

\affiliations {
    \textsuperscript{\rm 1}National and Kapodistrian University of Athens, Greece\\
    \textsuperscript{\rm 2}Archimedes, Athena Research Center, Greece\\
    \textsuperscript{\rm 3}The University of Edinburgh, UK\\
    \textsuperscript{\rm 4}Technical University of Denmark\\
    \textsuperscript{\rm 5}Pioneer Centre for AI, Denmark\\
    \textsuperscript{\rm 6}The University of Essex, UK\\
    \textsuperscript{\rm 7}Monzo Bank, UK
}

\usepackage{bibentry}
\begin{document}

\maketitle

% while also addressing the challenge of latent space control in LDMs.

\begin{abstract}
Adapting text-to-image (T2I) latent diffusion models (LDMs) to video editing has shown strong visual fidelity and controllability, but challenges remain in maintaining causal relationships inherent to the video data generating process. Edits affecting causally dependent attributes often generate unrealistic or misleading outcomes if these relationships are ignored. In this work, we introduce a causally faithful framework for counterfactual video generation, formulated as an Out-of-Distribution (OOD) prediction problem. We embed prior causal knowledge by encoding the relationships specified in a causal graph into text prompts and guide the generation process by optimizing these prompts using a vision-language model (VLM)-based textual loss. This loss encourages the latent space of the LDMs to capture OOD variations in the form of counterfactuals, effectively steering generation toward causally meaningful alternatives. The proposed framework, dubbed CSVC, is agnostic to the underlying video editing system and does not require access to its internal mechanisms or fine-tuning. We evaluate our approach using standard video quality metrics and counterfactual-specific criteria, such as causal effectiveness and minimality. Experimental results show that CSVC generates causally faithful video counterfactuals within the LDM distribution via prompt-based causal steering, achieving state-of-the-art causal effectiveness without compromising temporal consistency or visual quality on real-world facial videos. Due to its compatibility with any black-box video editing system, our framework has significant potential to generate realistic 'what if' hypothetical video scenarios in diverse areas such as digital media and healthcare. Code: \textcolor{blue}{\url{https://github.com/nysp78/counterfactual-video-generation}}.
\end{abstract}


\begin{figure}[ht]
  \centering
  \includegraphics[width=1.0\linewidth]{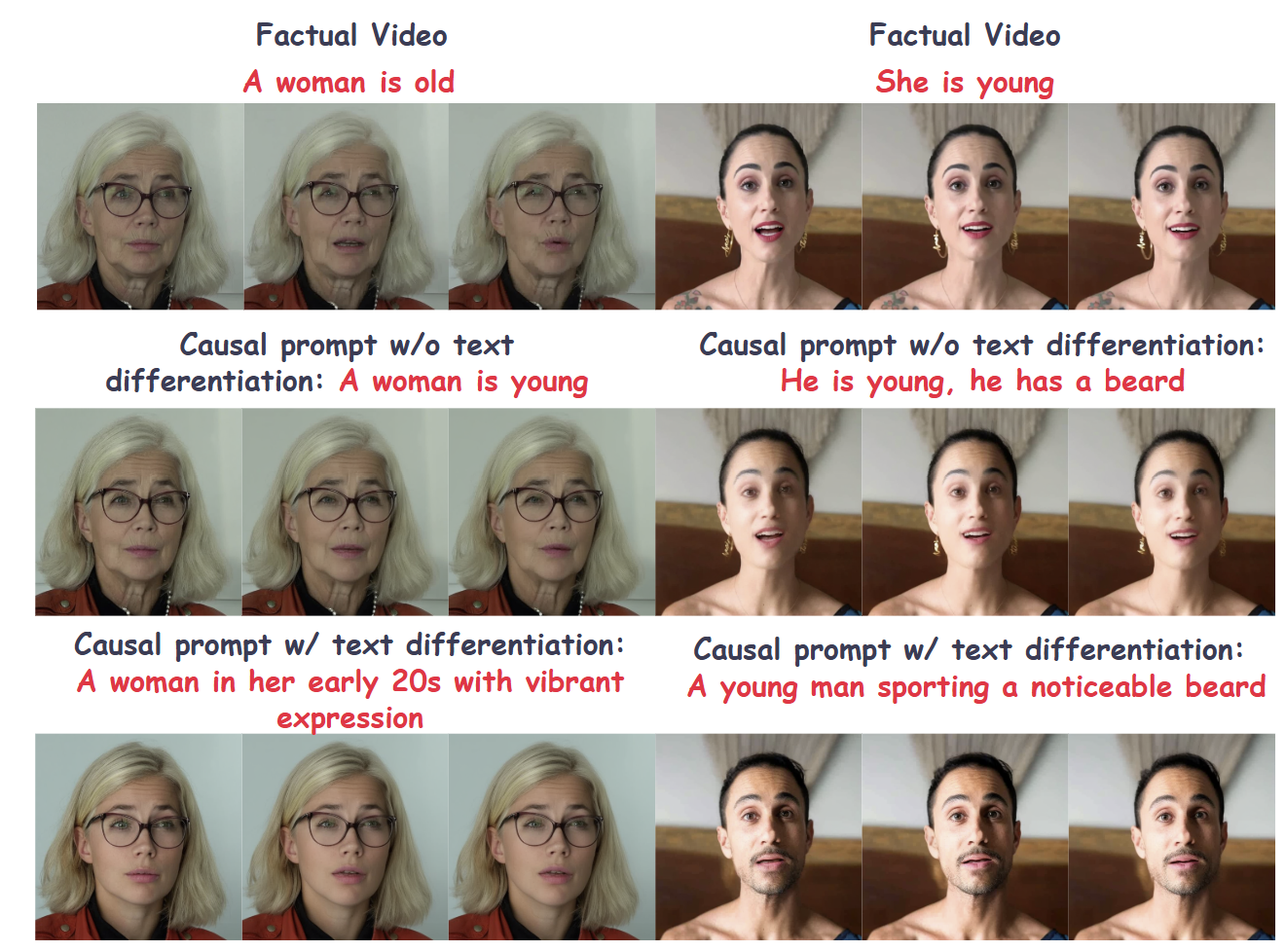}
  \caption{\textbf{Generated counterfactual results}: We intervene on age (make the woman young) and gender (transform a woman to a man with a beard). Our CSVC framework (3rd row) optimally steers the counterfactual generation process by causally tuning an initial target prompt achieving better results than w/o steering (2nd row).}
  \label{figure1_updated}
\end{figure}

\section{Introduction}
Text-to-image (T2I) latent diffusion models (LDMs) have significantly advanced the field of image generation~\cite{podell2024sdxl, rombach2022high}, showcasing remarkable fidelity and enhanced creative control in image editing~\cite{cong2024FLATTEN, feng2024ccedit, geyer2024tokenflow, jeong2024groundavideo, kara2024rave}. 
However, the efficacy of image editing is not consistent, as modifications affecting attributes with causal dependencies often generate unrealistic and potentially misleading results if these relationships are disregarded. This issue is particularly critical in data where causal interplays determine the imaging content~\cite{melistas2024benchmarking, pawlowski2020deep, papanastasiou2024confounder}.

Recent efforts in video editing adapt T2I models to address the challenge of maintaining spatiotemporal consistency~\cite{gu2024videoswap,liu2024video, shin2024edit, zhang2023towards, zhao2024motiondirector, cong2024FLATTEN, geyer2024tokenflow, wu2023tune}. While some of these techniques rely on adapting pre-trained models through a training or fine-tuning process to achieve text-guided video editing~\cite{gu2024videoswap, shin2024edit, zhang2023towards, zhao2024motiondirector}, other zero-shot~\cite{cong2024FLATTEN, geyer2024tokenflow} or one-shot~\cite{ wu2023tune, liu2024video} methods have focused on enabling text-driven video editing with a reduced reliance on extensive training. One- and zero-shot extensions of compact pre-trained LDMs democratize flexible video editing by removing heavy retraining and slashing compute costs.%presents challenges, primarily due to the need to maintain visual realism across temporal frames, which greatly increases the modeling complexity \cite{gu2024videoswap, liu2024video, shin2024edit, wu2023tune, zhang2023towards, zhao2024motiondirector}. 

Moreover, recent work highlights that suboptimal prompts can adversely affect outcomes in both video generation~\cite{ji2024prompt,cheng2025vpo,gao2025devil} and video editing~\cite{jeong2024groundavideo, liu2024video}, necessitating careful prompt consideration in these domains. This underscores the critical role of input text prompts in improving visual fidelity and semantic accuracy in video editing.

% \textcolor{red}{Yet, in contrast to these developments, informing video editing methods with prior causal knowledge through a predefined causal graph and performing controllable counterfactual generation within Pearl’s causality framework~\cite{pearl2009causality} has not been previously investigated in the video domain. We argue that prior causal knowledge can be incorporated into video editing systems via edited prompts designed to represent the relationships specified by the causal graph. This approach enables the generation of causally faithful video counterfactuals through controlled causal interactions and optimized causal prompts, which are crucial for achieving realistic video editing.} 
Yet, in contrast to these developments, existing video-editing methods overlook predefined causal graphs and Pearl-style \cite{pearl2009causality} counterfactual reasoning. We therefore propose encoding such graphs into prompt edits, enabling controlled interventions on targeted attributes and producing causally faithful and realistic video counterfactuals without additional training. Our approach leverages a vision-language model (VLM) loss to guide the latent space of the diffusion model toward generating Out-of-Distribution (OOD) samples that reflect counterfactual scenarios aligned with the encoded causal structure.
%Causally faithful video counterfactuals via controlled causal interactions and optimized prompts, are essential for realistic video editing.

In addition, LDMs have demonstrated tremendous capabilities in image and video editing~\cite{huang2025diffusion, sun2024diffusion}; however, their intricate latent spaces present challenges for direct manipulation and effective control during the generation process \cite{park2023understanding, kwon2022diffusion}. Inspired by prompt optimization for black-box LDMs \cite{hao2023optimizing, manas2024improving}, we posit that text prompt modification offers an implicit yet powerful way to steer generation towards effective, realistic counterfactual estimations. We hypothesize that optimizing causally consistent prompts is key to controlling causal consistency and achieving effective, realistic video counterfactuals.

% \textcolor{red}{We approach LDM-based video editing as the generation of video counterfactuals, framing it as a specific instance of Out-of-Distribution (OOD) generation \cite{pearl2009causality, scholkopf2021towardcrl, ribeiro2023high}, where the goal is to modify specific attributes of a factual (source) video and produce samples that differ from the original training distribution while remaining causally consistent and realistic. To generate plausible and semantically meaningful video counterfactuals, we introduce a novel framework that encodes a prior causal graph into text prompts and combines it with a vision-language model (VLM) loss using textual differentiation~\cite{textgrad5optimizing} to achieve optimized, prompt-driven causal steering. Both the VLM and the LDM are treated as black boxes, allowing us to focus on their interaction without the need for explicit manipulation or particular knowledge of their internal workings. Figure \ref{figure1_updated} depicts how counterfactual estimations improve with a causally consistent prompt using text differentiation optimization~\cite{textgrad5optimizing}}.%, compared to one without. 

We formulate LDM-based video editing as a task of generating video counterfactuals, conceptualized as a structured instance of OOD generation~\cite{pearl2009causality, scholkopf2021towardcrl, ribeiro2023high}. The objective is to modify specific attributes of a factual (source) video while ensuring the edited samples remain semantically coherent and causally consistent with respect to a predefined causal structure. To this end, we propose a novel framework that encodes prior causal knowledge into natural language prompts, integrating them with a VLM through a differentiable textual contrast loss~\cite{textgrad5optimizing}. This loss facilitates prompt optimization via textual differentiation, guiding the diffusion process toward causally faithful edits. Crucially, both the VLM and LDM are treated as black-box components, enabling general applicability without requiring architectural modifications or internal access. As illustrated in Figure~\ref{figure1_updated}, counterfactual fidelity improves significantly when prompts are optimized to reflect causal constraints using our text-guided differentiation approach, effectively shaping the LDM latent space to generate OOD counterfactuals that align with the intended interventions.

Our methodology addresses the challenge of explicitly guiding the high-dimensional latent space of LDMs to achieve specific, targeted OOD counterfactual modifications. By manipulating the input text prompt with causal guidance, our approach steers the LDM's transformations during inference toward the desired OOD counterfactual outcome. This process allows for human-controllable prompt tuning, enabling the generation of causally consistent counterfactuals. The VLM counterfactual loss–optimized text conditioning directs the LDM-based editing system, ensuring that the generated video frames align with the desired counterfactual changes in a causally consistent manner, thereby effectively controlling the generation of diverse counterfactuals.

%directs the denoising process at each diffusion timestep

This paper proposes a novel framework "\textbf{C}ausal \textbf{S}teering for \textbf{V}ideo \textbf{C}ounterfactuals" (CSVC) which, to the best of our knowledge, is the first to integrate causal priors into video editing systems by encoding predefined causal relationships into text prompts. To generate causally faithful counterfactuals, we propagate textual gradients \cite{textgrad5optimizing} through these prompts using a VLM-based loss, guiding the diffusion model to produce semantically meaningful OOD edits aligned with the intended interventions.

% prompt automate the creation of causally faithful, well-structured, and model-aligned textual prompts,
 
In summary, our contributions are: 

%Consequently, the main focus of this work is to automate the creation of causally faithful, well-structured, and model-aligned textual prompts, steering the counterfactual transformations toward accurate and semantically meaningful OOD edits at inference. In summary, our contributions are: 

% \textbf{Main contributions}:
    \begin{itemize}
   \item We propose the first causal steering framework (CSVC) for diffusion-based video editing, enabling the generation of causally faithful OOD counterfactuals by optimizing text prompts using VLM feedback.

    \item CSVC operates entirely in a black-box setting, requiring no access to the internals of the video editing system or the LDM backbone, making it broadly applicable across zero- and one-shot video editing methods.

    \item We introduce a principled optimization strategy that refines causal prompts via textual gradients, steering the latent space of LDMs toward semantically meaningful and causally aligned counterfactuals.

    \item Our approach achieves state-of-the-art causal effectiveness on diverse real-world facial videos across multiple interventions (e.g., age, gender, beard, baldness) while preserving video quality, minimality, and temporal coherence.

    \item We design novel VLM-based metrics to assess causal effectiveness and minimality, offering interpretable and scalable evaluation tools for counterfactual video generation.

%\item We incorporate prior causal knowledge into diffusion-based video editing systems by providing  prompts that integrate causal relationships of a predefined causal graph.

%\item To the best of our knowledge, we are the first to propose a novel framework that allows steering diffusion-based video editing towards causal OOD counterfactuals by propagating textual feedback from a VLM-based counterfactual loss through the LDM input prompt.

%\item Experimentally, we show that our framework improves the causal effectiveness of counterfactual estimations on real-world diverse facial videos by tuning the input prompt without accessing LDM internals, while preserving video quality, minimality, and temporal consistency. We also demonstrate that causally faithful steering enables generating consistent counterfactuals from the LDM latent space.

%\item We demonstrate that causally faithful steering enables causally faithful counterfactual generation from LDM latent spaces.

%\item We design VLM-based evaluation metrics to further assess the capacity of diffusion-based video editing frameworks for plausible counterfactual generation.

    \end{itemize}

\begin{figure*}[ht]
  \centering
  \includegraphics[width=0.82\linewidth]{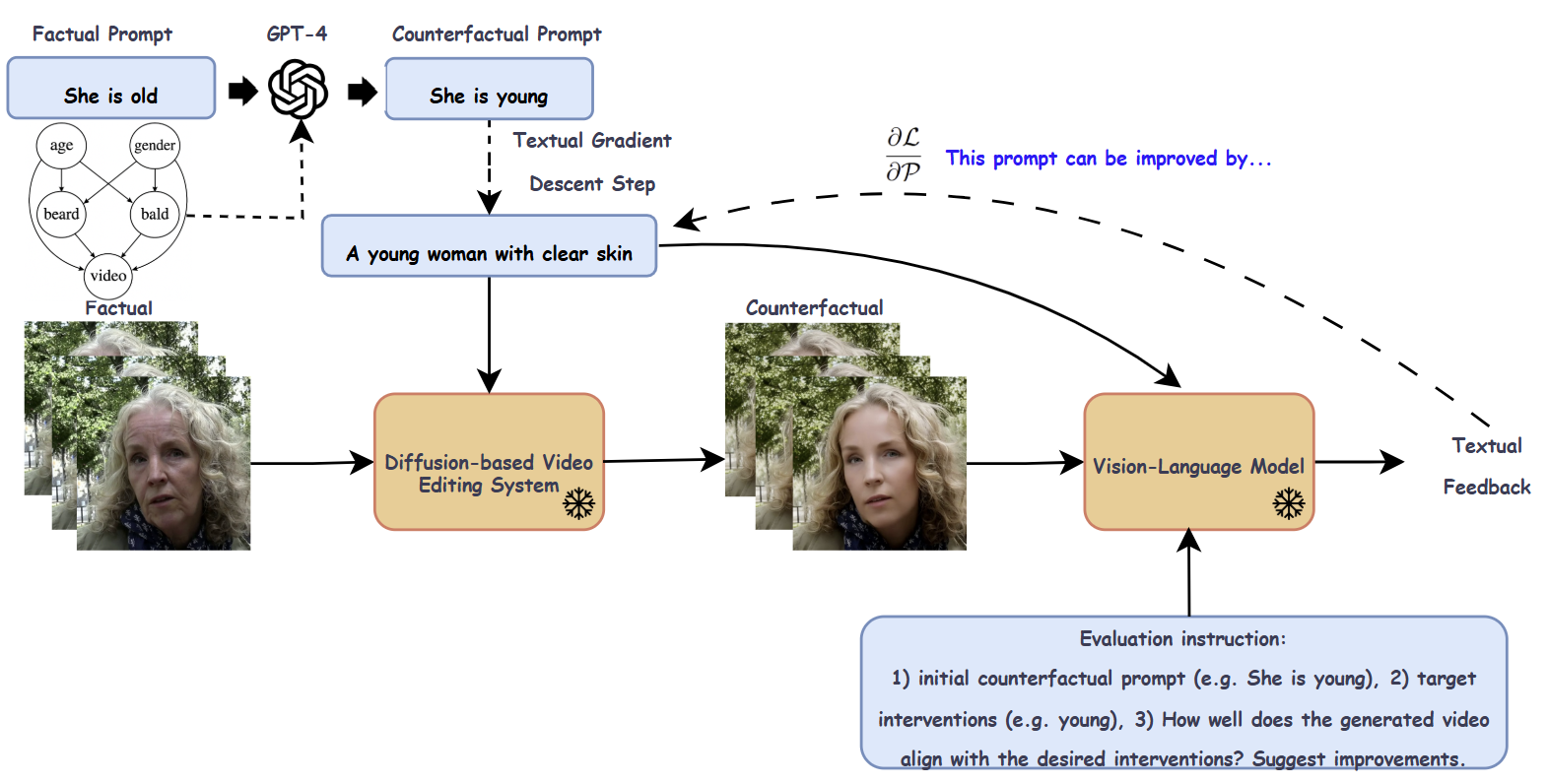}
  \caption{\textbf{CSVC at a glance:} The initial counterfactual prompts (e.g., She is young) are generated using GPT-4 by providing the causal graph and the factual prompts (e.g., She is old) and leveraging in-context learning \cite{dong2022survey_incontext}. The video editing system operates as a black-box (frozen) counterfactual generator and the (black-box) VLM as an evaluator of the generated counterfactuals. The VLM takes as input a generated counterfactual frame, the evaluation instruction, and the target counterfactual prompt $\mathcal{P}$, and outputs textual feedback used to compute a ‘textual gradient’ $\frac{\partial \mathcal{L}}{\partial \mathcal{P}}$, which guides the optimization of $\mathcal{P}$ by focusing on the unsuccessful interventions.
}
  \label{fig:figure_2}
\end{figure*}

\section{Related Work}
 \noindent\textbf{Latent Diffusion-based Video Editing.} LDMs~\cite{podell2024sdxl, rombach2022high} have significantly advanced video generation and editing~\cite{croitoru2023diffusion, sun2024diffusion}. Tuning-based methods focus on either adapting text-to-image models \cite{podell2023sdxl} through cross-frame attention and one-shot tuning \cite{zhang2023towards, wu2023tune, liu2024video, shin2024edit, gu2024videoswap}, or on fine-tuning text-to-video models with one- or multi-shot tuning~\cite{wang2025videodirector, zhao2024motiondirector}. Controlled editing methods, like ControlNet \cite{chen2023control}, use priors such as optical flow~\cite{yang2023rerender, hu2023videocontrolnet}, depth maps \cite{feng2024ccedit}, or pose information~\cite{ma2024follow, yang2025videograin} to enforce consistency. Training-free methods use diffusion features~\cite{tang2023emergent}, latent fusion~\cite{qi2023fatezero, khandelwal2023infusion}, noise shuffling~\cite{kara2024rave}, or optical-flow guidance~\cite{chu2024medm, cong2024FLATTEN, yang2024fresco, jeong2024groundavideo}. Leveraging lightweight one and zero-shot T2I LDM-based video editing, we investigate how prompt optimization with causal priors and text differentiation enables causal steering to generate effective, minimal, high-quality, and temporally consistent video counterfactuals.

 %~\cite{sanchez2022diffusion, melistas2024benchmarking}.
 
\noindent\textbf{Counterfactual Image and Video Generation.} Visual counterfactual generation explore hypothetical ``what-if'' scenarios through targeted and semantically meaningful modifications to the input~\cite{wachter2017counterfactual, scholkopf2021toward}. It is applied in counterfactual explainability ~\cite{verma2024counterfactual,augustin2022diffusion,jeanneret2022diffusion,jeanneret2023adversarial,weng2024fast,pegios2024diffusion,pegios2024counterfactual,sobieski2025rethinking}, robustness testing ~\cite{dash2022evaluating, prabhu2023lance,le2023coco,lai2024improving,yu2024revisiting,zhang2024learning,weng2024fast}, and~causal inference \cite{pearl2009causality, vlontzos2022review,vlontzos2023estimating,VLONTZOS2025367,pawlowski2020deep,kocaoglu2018causalgan,xia2021learning,abdulaal2022deep,sanchez2022diffusion,ribeiro2023high,sanchez2022healthy,fontanella2024diffusion,song2024doubly}. While much work focuses on static images~\cite{monteiromeasuring, ribeiro2023high, melistas2024benchmarking}, the temporal coherence of causal counterfactual video generation remains underexplored~\cite{reynaud2022d}. In our CSVC framework, we integrate causal relationships and text-differentiation-based prompt optimization into three LDM-based methods via a VLM counterfactual loss to generate effective video counterfactuals.

\noindent\textbf{Evaluation of Visual Editing and Counterfactuals.}  %Evaluating counterfactuals is challenging \cite{scholkopf2021toward, melistas2024benchmarking}. While standard metrics assess image quality \cite{korhonen2012peak, zhang2018unreasonable, wang2004image, heusel2017gans} and semantic alignment \cite{radford2021learning}, causal counterfactuals~\cite{melistas2024benchmarking,galles1998axiomatic,halpern2000axiomatizing} require stricter criteria like causal effectiveness \cite{monteiromeasuring} and minimality \cite{sanchez2022diffusion}. In video, evaluation is more complex due to the temporal consistency required. Existing video benchmarks \cite{liu2023fetv, yuan2024chronomagic, liu2024evalcrafter, huang2024vbench, ji2024t2vbench,sun2024diffusion} overlook counterfactual reasoning. In addition, commonly used metrics in video generation such as DOVER \cite{wu2023dover}, CLIP Score \cite{radford2021learning}, and flow warping error \cite{lai2018learning} do not assess causal relationships. We evaluate the generated counterfactual videos both in terms of causal adherence through counterfactual effectiveness and minimality \cite{monteiromeasuring, ribeiro2023high, melistas2024benchmarking} and in terms of general video quality and temporal consistency. For minimality, we introduce a novel metric based on vision-language models. This comprehensive evaluation allows us to thoroughly assess causal adherence and the quality of counterfactual generation in text-guided video editing.
%We evaluate our methodology on a subset of the CelebV-Text data \cite{yu2023celebv}. 
Evaluating counterfactuals is inherently challenging \cite{scholkopf2021toward, melistas2024benchmarking}. Standard metrics assess image quality \cite{korhonen2012peak, zhang2018unreasonable, wang2004image, heusel2017gans} and semantic alignment \cite{radford2021learning}, but causal counterfactuals \cite{melistas2024benchmarking, galles1998axiomatic, halpern2000axiomatizing} require stricter criteria, such as causal effectiveness \cite{monteiromeasuring} and minimality \cite{sanchez2022diffusion}. In video, evaluation is further complicated by the need for temporal consistency, while existing benchmarks \cite{liu2023fetv, yuan2024chronomagic, liu2024evalcrafter, huang2024vbench, ji2024t2vbench, sun2024diffusion} largely overlook counterfactual reasoning. Additionally, widely used video metrics such as DOVER \cite{wu2023dover}, CLIP Score \cite{radford2021learning}, and flow warping error \cite{lai2018learning} fail to capture causal relationships. To address this, we evaluate generated counterfactual videos using both causal adherence--via counterfactual effectiveness and minimality \cite{monteiromeasuring, ribeiro2023high, melistas2024benchmarking}--and overall video quality and temporal consistency. For minimality, we introduce a novel VLM-based metric, enabling comprehensive assessment of causal fidelity in text-guided video counterfactual generation.

\section{Background}
\textbf{T2I LDMs for Video Editing.} Recent text-guided video editing methods \cite{wu2023tune, cong2024FLATTEN, geyer2024tokenflow} 
%such as Tune-A-Video, FLATTEN and TokeFlow , 
employ pre-trained T2I LDMs, typically Stable Diffusion \cite{rombach2022high}, that operate on a latent image space. A pre-trained autoencoder $(\mathcal{E}, \mathcal{D})$~\cite{kingma2013autoVAE,van2017neuralVQ-VAE} maps an image frame $x$ to a latent code $z=\mathcal{E}(x)$, with $\mathcal{D}(z) \approx x$.  A conditional U-Net~\cite{ronneberger2015u} denoiser $\epsilon_\theta$ is trained to predict noise in the latent $z_t$ at diffusion timestep $t$, minimizing: \begin{equation*}
E_{z,\epsilon\sim\mathcal{N}(0,1),\,t,\,c}\!\bigl[\|\epsilon - \epsilon_\theta(z_t,t,c)\|_2^2\bigr]
\end{equation*}
,
where %c=\psi(\mathcal{P})$ 
$c$
is the 
%CLIP
embedding of text prompt $\mathcal{P}$.
The U-Net $\epsilon_\theta$ can be either inflated into a 3D spatio-temporal network for one-shot video fine-tuning~\cite{wu2023tune} and zero-shot optical-flow guidance~\cite{cong2024FLATTEN}, or directly used for frame editing, with temporal consistency imposed via feature propagation~\cite{geyer2024tokenflow}. These methods leverage deterministic DDIM~\cite{song2020denoisingDDIM} sampling and inversion which allows to reconstruct or edit the original video frames. Although each method has its own temporal regularization strategies and heuristics, given an input video 
$\mathcal{V}$ and an editing prompt $\mathcal{P}$, the core video editing process can be expressed as:
\begin{equation}
\mathcal{V'} = \mathcal{D}(DDIM-samp(DDIM-inv(\mathcal{E}(\mathcal{V})), \mathcal{P}))
\label{eq:framework}
\end{equation}
\textbf{Causal Framework for Video Editing.}
Within Pearl’s counterfactual inference paradigm \cite{pearl2009causality} of abduction–action–prediction, DDIM inversion corresponds to the \emph{abduction} step, the \emph{action} step is the prompt-based intervention using the editing prompt $\mathcal{P}$, and DDIM sampling performs the \emph{prediction}, producing the  counterfactual video $\mathcal{V'}$.

\section{Methodology}
\label{Method}
\subsection{Causal Graph Integration}
Prior causal knowledge can be injected into video editing systems via target prompts that encode the causal relationships defined by a DAG. As shown in Figure \ref{fig:figure_2}, we provide GPT-4 with a causal DAG and factual (source) prompts and, using \emph{in-context learning} \cite{dong2022survey_incontext}\footnote{Details on the in-context learning prompt are provided in the Appendix.}, generate causally consistent OOD counterfactual prompts representing interventions aligned with the predefined causal relationships.

\subsection{Causal Steering for Video Counterfactuals (CSVC)}
\label{VLM-base_loss}
\textbf{Video Editing System as a Counterfactual Generator.} 
We treat the video editing method as an opaque black-box system for counterfactual generation (Figure~\ref{fig:figure_2}), assuming no access to the \(\epsilon_\theta\) LDM parameters (no updates or backpropagation) and no control over internal mechanisms such as DDIM sampling or inversion. For any prompt-based video editing system \(f\), an input video \(\mathcal{V}\) and a counterfactual (editing) prompt \(\mathcal{P}\), Equation~\ref{eq:framework} simply becomes: $\mathcal{V'} = f(\mathcal{V}, \mathcal{P})$. Our CSVC framework is compatible with any black-box, text-guided diffusion video editing system and is evaluated with three such methods. Since the causal counterfactual prompt \(\mathcal{P}\) critically impacts the counterfactual output \(\mathcal{V'}\)~\cite{hao2023optimizing,ji2024prompt,jeong2024groundavideo}, we further refine \(\mathcal{P}\) using \emph{textual feedback} from an external optimizer~\cite{textgrad5optimizing}.

\textbf{VLM-based counterfactual loss for steering the video generation.} To enable the generation of OOD video counterfactuals, we integrate the causal counterfactual prompts into the video editing system and design a VLM-based loss to encourage the LDM backbone to generalize to unobserved contexts by causally optimizing these input prompts. Suboptimal prompts can degrade video editing quality, making effective prompt refinement crucial~\cite{hao2023optimizing,mo2024dynamic,manas2024improving,ji2024prompt,cheng2025vpo,gao2025devil,jeong2024groundavideo,liu2024video}. Manual prompt engineering~\cite{liu2022design} and simple paraphrasing~\cite{gao2021making,haviv2021bertese} offer partial solutions, while black-box optimization methods typically fine-tune a large language model (LLM) as a model-specific prompt interface~\cite{hao2023optimizing,mo2024dynamic,ji2024prompt,cheng2025vpo}. Others explore prompt paraphrases by iteratively updating in-context examples~\cite{manas2024improving}. To automate counterfactual generation for any text-guided video editing system, we employ TextGrad~\cite{textgrad5optimizing} which enables prompt-level causal steering by optimizing counterfactual prompts based on the underlying causal relationships and target interventions. TextGrad leverages LLMs to generate natural-language ``textual gradients'' used for iterative refinement of complex systems through textual feedback. Building on this, we design a  counterfactual ``multimodal loss'' using a VLM to guide the video generation towards the target interventions. The proposed CSVC framework is illustrated in Figure \ref{fig:figure_2}.

Given a generated counterfactual video frame, the counterfactual prompt, and an evaluation instruction containing the target interventions, we implement our proposed ``multimodal loss'' using a VLM:
\begin{equation}
\mathcal{L} = \textit{VLM}(\mathcal{V'_{\mathit{frame}}}, \textit{evaluation instruction}, \mathcal{P}),
\label{eq:loss}
\end{equation}
where the \emph{evaluation instruction} \footnote{Prompt details are available in the appendix.} is a well-defined textual input to the VLM to suggest improvements on $\mathcal{P}$ based on how well the generated visual input $\mathcal{V}'_{\mathit{frame}}$ (extracted from $\mathcal{V'}$) aligns with the target counterfactual interventions. We further augment the \emph{evaluation instruction} with a \emph{causal decoupling}\footnote{A causal decoupling example prompt is provided in the appendix.} text input that instructs the VLM to ignore \emph{upstream} variables when intervening on \emph{downstream} ones. This yields optimized prompts that omit explicit upstream references (e.g., neutralizing gender), enabling the LDM backbone to generate samples that intentionally violate the causal graph, such as rendering a woman with a beard (Figure \ref{fig:figure5}).

To optimize $\mathcal{P}$, we employ \emph{Textual Gradient Descent} (TGD)~\cite{textgrad5optimizing}, which directly updates the prompt:
\begin{equation}
\begin{aligned}
\mathcal{P}' &= \text{TGD.step}\!\left(\mathcal{P}, \frac{\partial \mathcal{L}}{\partial \mathcal{P}}\right) \\
& =\textit{LLM}\Big(
  \textit{Below are the criticisms on }\{\mathcal{P}\}: 
  \left\{\frac{\partial \mathcal{L}}{\partial \mathcal{P}}\right\}, \\
&\quad \textit{Incorporate the criticisms and produce a new prompt.}
\Big)
\end{aligned}
\label{TGD_update}
\end{equation}

where \emph{$\frac{\partial \mathcal{L}}{\partial \mathcal{P}}$} \footnote{Due to space constraints, we encourage the interested reader to refer to the Appendix for an explanation of the textual gradients computation.} denotes the ``textual gradients'', passed through an \textit{LLM} \footnote{For simplicity and robustness, we employ the same LLM/VLM model (GPT‑4) for all operations.} at each TGD update to generate a new prompt incorporating the VLM criticisms. % i.e, the suggested textual feedback from Equation~\ref{eq:loss}.
%In each TGD update, an LLM takes as input the ``textual gradients'' --criticisms from the VLM loss  in Equation~\ref{eq:loss} on how to improve $\mathcal{P}$-- and generates a new prompt $\mathcal{P}'$.
Optimization halts when the target interventions are met % i.e, LLM is instructed to return ``no optimization needed'' 
or the maximum number of iterations is reached. A summary of the proposed CSVC approach is showcased in Algorithm 1.

%\textbf{Overall algorithm}
\begin{algorithm}[tb]
\caption{Causal Steering for Video Counterfactuals}
\label{alg:counterfactual}
\textbf{Input}: Causal Counterfactual prompt $\mathcal{P}$, Factual video $\mathcal{V}$, DiffusionVideoEditor, \textit{VLM}, \textit{evaluation instruction}\\
\textbf{Parameter}: Maximum iterations $\mathit{maxIters}$\\
\textbf{Output}: Counterfactual video $\mathcal{V'}$
\begin{algorithmic}[1]

\STATE $prompt \gets \mathcal{P}$ \COMMENT{Initialize prompt}
\STATE $optimizer \gets \mathrm{TGD}(\mathrm{parameters}=[prompt])$ \COMMENT{Set up textual optimizer}
%{\color{blue} \qquad  Set up textual optimizer}

\FOR{$iter = 1  \; \TO \;  \mathit{maxIters}$}
    \STATE $\mathcal{V'} \gets \mathrm{DiffusionVideoEditor}(\mathcal{V}, prompt)$ \COMMENT{Generate counterfactual (Eq.~\ref{eq:framework})}
    \STATE $loss \gets \textit{VLM}(\mathcal{V'_{\mathit{frame}}}, \textit{evaluation instruction}, prompt)$ \COMMENT{Evaluate (Eq.~\ref{eq:loss})}

    \IF{$\text{``no optimization''} \in loss.value$}
        \STATE \textbf{break}
    \ENDIF

    \STATE $loss.\mathrm{backward}()$ \COMMENT{Compute $\frac{\partial \mathcal{L}}{\partial \mathcal{P}}$}
    \STATE $optimizer.\mathrm{step}()$ \COMMENT{Update prompt via TGD (Eq.~\ref{TGD_update})}
\ENDFOR

\STATE \textbf{return} Final counterfactual video $\mathcal{V'}$

\end{algorithmic}
\end{algorithm}

\subsection{VLMs for assessing causal effectiveness}
\label{para_3.4_effectiveness_VLM}

Effectiveness is key in counterfactual generation, indicating if the target intervention succeeded~\cite{galles1998axiomatic, monteiromeasuring, melistas2024benchmarking}. CLIP-based metrics \cite{radford2021learningCLIP} lack interpretability and are inefficient for capturing \emph{causal} alignment between text and image. Following~\citep{hu2023TIFA}, we use a VLM to assess effectiveness across a set of generated counterfactual videos with a visual question answering (VQA) approach. Given triplets $\{Q^{\alpha}_{i}, C_i,  V'_{\mathit{frame_{i}}}\}^{N}_{i=1}$, where $Q^{\alpha}_{i}$ is a multiple-choice question about the intervened attribute $a$, $C_i$ is the correct answer extracted from the target counterfactual prompt, and $\mathcal{V'_{\mathit{frame_{i}}}}$ is a generated counterfactual video frame, we measure effectiveness by the accuracy of the VLM's answer:
\begin{equation}
\mathit{Effectiveness}(\alpha) = \frac{1}{N} \sum_{i=1}^{N} 1 \left[ \textit{VLM}(\mathcal{V'_{\mathit{frame_{i}}}}, Q^{\alpha}_{i}) = C_i \right].
\label{eq:effectiveness}
\end{equation}

\subsection{VLMs for assessing minimality}
\label{para_3.5_minimality_VLM}
Minimal interventions~\cite{scholkopf2021toward, sanchez2022diffusion, melistas2024benchmarking} are considered a principal property for visual counterfactuals. In counterfactual generation a substantial challenge lies in incorporating the desired interventions (edits), while preserving unmodified other visual factors of variation which are not related to the assumed causal graph~\cite{monteiromeasuring} -- a challenge closely tied to identity preservation of the observation (factual) \cite{ribeiro2023high}. We evaluate counterfactual minimality in the text domain, offering a more interpretable alternative to conventional image-space metrics~\cite{zhang2018unreasonableLPIPS}. Specifically, we prompt a VLM to describe in detail both factual and counterfactual frames, excluding attributes associated with the assumed causal graph. We then embed the resulting descriptions using a BERT-based sentence transformer~\cite{wang2020minilm} and compute their cosine similarity in the semantic space.  The overall minimality metric can be expressed as follows: 

{\small
\begin{equation*}
\mathcal{P}_{min} = \text{"Describe this frame in detail, exclude DAG variables"}
\end{equation*}
}

\begin{equation}
\begin{split}
Min(\mathcal{V_{\mathit{frame}}}, \mathcal{V'_{\mathit{frame}}}) =
\cos\big(&\tau_\phi(\textit{VLM}(\mathcal{V_{\mathit{frame}}}, \mathcal{P}_{min})), \\
         &\tau_\phi(\textit{VLM}(\mathcal{V'_{\mathit{frame}}}, \mathcal{P}_{min}))\big)
\end{split}
\label{eq:minimality}
\end{equation}

where $\tau_\phi(.)$ denotes the semantic text encoder and $V_{frame}, V'_{frame}$ the factual and counterfactual frames.

\section{Experiments and Results}
\label{Exps}

\subsection{Experimental Setup}

%\begin{figure}[t]
%  \centering
%  \includegraphics[width=0.15\textwidth]{figures/causal_graph_v2.png}
%  \caption{CelebV-Text causal graph.}
%  \label{fig:causal_graph}
%\end{figure}
 %These counterfactual prompts were either contributed by the authors or generated by ChatGPT, based on the causal relationships defined by the assumed causal graph.
\begin{table*}[t]
\centering
\scriptsize
\setlength{\tabcolsep}{5.9pt} % reduce spacing for all columns
\renewcommand{\arraystretch}{0.95}

\begin{tabular}{lcccc||ccccc}
\toprule
\textbf{Method} &
\multicolumn{4}{c}{\textbf{Effectiveness (VLM Acc.)}} &
\multicolumn{2}{c}{\textbf{Minimality}} &
\multicolumn{3}{c}{\textbf{Video Quality \& Temp. Consistency}} \\
\cmidrule(lr){2-5}\cmidrule(lr){6-7}\cmidrule(l){8-10}
 & \texttt{age} $\uparrow$ & \texttt{gender} $\uparrow$ & \texttt{beard} $\uparrow$ & \texttt{bald} $\uparrow$
 & LPIPS $\downarrow$ & VLM-Min $\uparrow$
 & DOVER $\uparrow$ & FVD ($\times 10^{-2}$) $\downarrow$ & CLIP-Temp $\uparrow$ \\
\midrule

\textbf{FLATTEN}  \\
Initial Prompt & 0.597 & 0.746 & 0.313 & 0.418 & \textbf{0.161} & \textbf{0.791} & \textbf{0.841} & \textbf{3.472} & \textbf{0.982} \\
LLM Paraphrasing & 0.582 & 0.791 & 0.299 & 0.179 & 0.178 & 0.786 & \textbf{0.841} & 3.662 & \textbf{0.982} \\
\rowcolor{gray!20}
CSVC w/o causal decoupling & 0.701 & 0.791 & 0.343 & 0.403 & 0.179 & 0.789 & 0.828 & 4.162 & 0.981 \\
\rowcolor{gray!20}
CSVC w/ causal decoupling & \textbf{0.731} & \textbf{0.806} & \textbf{0.582} & \textbf{0.433} & 0.179 & 0.781 & 0.834 & 4.188 & \textbf{0.982} \\
\midrule

\textbf{Tune-A-Video} \\
Initial Prompt & 0.529 & \textbf{0.985} & 0.412 & 0.824 & \textbf{0.320} & \textbf{0.742} & 0.557 &  \textbf{9.814} & \textbf{0.956} \\
LLM Paraphrasing & 0.507 & 0.970 & 0.433 & 0.358 & 0.396 & 0.695 & \textbf{0.596} & 13.581 & 0.939 \\
\rowcolor{gray!20}
CSVC w/o causal decoupling & 0.779 & \textbf{0.985} & 0.426 & 0.868 & 0.362 & 0.722 & 0.552 & 11.600 & 0.955 \\
\rowcolor{gray!20}
CSVC w/ causal decoupling & \textbf{0.824} & \textbf{0.985} & \textbf{0.676} & \textbf{0.912} & 0.370 & 0.717 & 0.558 & 11.840 & 0.955 \\
\midrule

\textbf{TokenFlow} \\
Initial Prompt & 0.672 & 0.836 & 0.388 & 0.522 & \textbf{0.227} & \textbf{0.776} & 0.787 & 7.712 & 0.984 \\
LLM Paraphrasing & 0.627 & 0.910 & 0.328 & 0.194 & 0.244 & 0.766 & \textbf{0.797} & \textbf{7.353} & 0.983 \\
\rowcolor{gray!20}
CSVC w/o causal decoupling & 0.909 & \textbf{0.925} & 0.426 & 0.552 & 0.241 & 0.773 & 0.784 & 8.060 & 0.984 \\
\rowcolor{gray!20}
CSVC w/ causal decoupling & \textbf{0.940} & 0.910 & \textbf{0.761} & \textbf{0.701} & 0.253 & 0.768 & 0.786 & 8.660 & \textbf{0.986} \\
\bottomrule
\end{tabular}
\caption{Counterfactual Evaluation: Effectiveness, Minimality, Video Quality \& Temporal Consistency. Highlighted rows show CSVC with and without the causal decoupling prompt.}

\label{all_metrics}
\end{table*}

\subsubsection{Evaluation Dataset.} 
Following standard video editing evaluation protocols~\cite{wu2023tune, geyer2024tokenflow, cong2024FLATTEN, liu2024video, qi2023fatezero, ku2024anyvv, wang2025videodirector}, we curated 67 text–video pairs from CelebV-Text~\cite{celebv-text}, an in-the-wild facial video dataset. For each video, we used the first 24 frames resized to 512$\times$512 and assumed the data-generating process follows the causal graph in Figure~\ref{fig:figure_2} \cite{yang2020causalvae, melistas2024benchmarking, kladny2023deepbacktracking}. Using GPT-4 with in-context learning, we generated four counterfactual prompts per source prompt, intervening on ‘age,’ ‘gender,’ ‘beard,’ and ‘baldness’ (Figure\ref{fig:figure_2}). For each edited prompt, we created four multiple-choice questions, each targeting a variable from the causal graph, to evaluate causal effectiveness using the VLM (Equation \ref{eq:effectiveness}). Additional details about the dataset are provided in the Appendix.
 \label{evaluation_dataset}

\subsubsection{Implementation Details.} We evaluate our CSVC framework with three diffusion-based video editing systems that perform zero- or one-shot editing by adapting T2I LDMs for efficient video manipulation. FLATTEN \cite{cong2024FLATTEN} uses optical flow–guided attention to enhance temporal coherence, Tune-A-Video \cite{wu2023tune} fine-tunes spatio-temporal attention on a single text-video pair, and TokenFlow \cite{geyer2024tokenflow} applies an image editing method to a set of keyframes and then propagates the edits to the remaining frames. We select these methods for their resource efficiency (FLATTEN: zero-shot; Tune-A-Video: one-shot; TokenFlow: zero-shot). Cross-attention-guided approaches such as Video-P2P \cite{liu2024video} or FateZero \cite{qi2023fatezero} are excluded as they require identical source and edited prompt structures. For consistency, all methods use Stable Diffusion v2.1 as the backbone. We adopt DDIM sampling with 50 steps and classifier-free guidance~\cite{ho2021classifierfree} (scale 4.5 for Tune-A-Video and TokenFlow, 7.5 for FLATTEN). The VLM counterfactual loss (Equation \ref{eq:loss}) is implemented with GPT-4 and optimized via TextGrad \cite{textgrad5optimizing} for 2 TGD iterations. For the VLM effectiveness metric (Equation \ref{eq:effectiveness}), we use LLaVA-NeXT\cite{li2024llava-next_}, and for minimality (Equation \ref{eq:minimality}) we use GPT-4 \cite{achiam2023gpt}, which can generate descriptions excluding causal graph variables, making it well-suited for our minimality metric. All experiments were run on a single A100 GPU.

\subsection{Results}

\begin{figure*}[ht]
  \centering
  \includegraphics[width=0.8\linewidth]{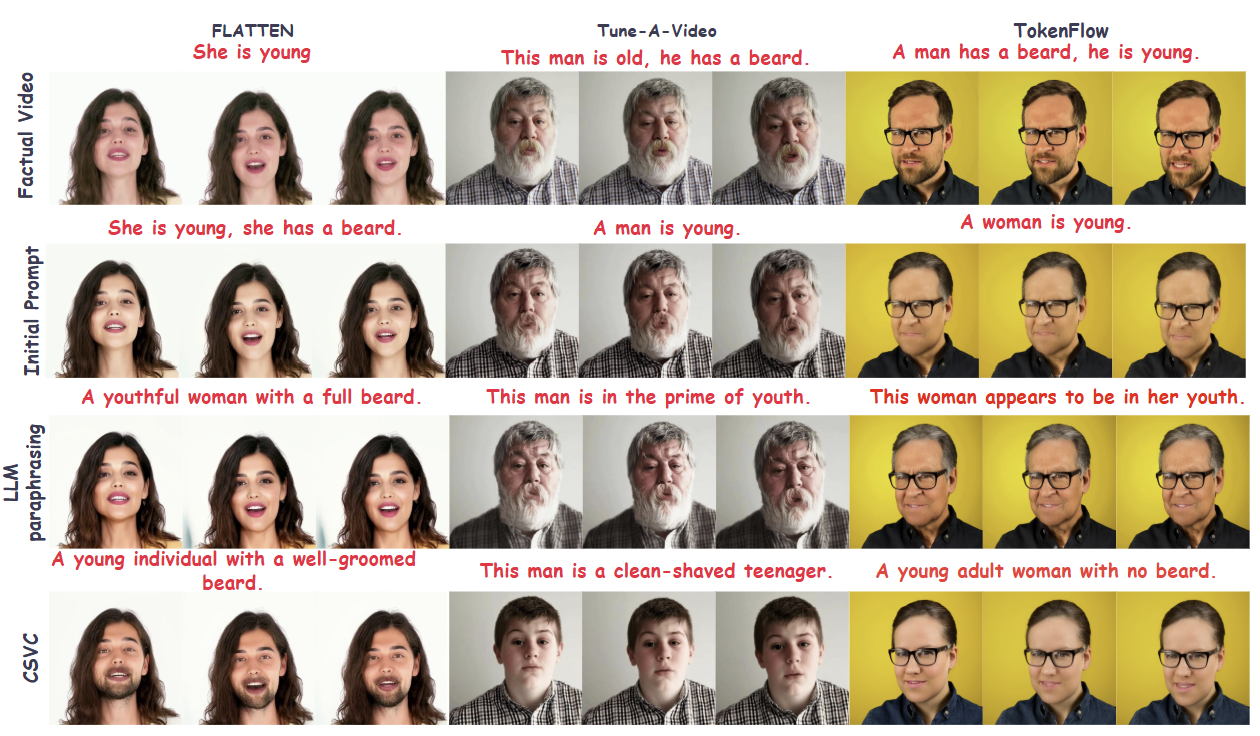}
\caption{\textbf{Qualitative results}: \textbf{First panel:} intervention on beard (adding a beard to a woman: breaking strong causal dependencies). \textbf{Second panel:} intervention on age (making an old man with a beard appear young with no beard). \textbf{Third panel:} intervention on gender (transforming a man with a beard into a woman). The accuracy of the edits in the bottom row demonstrates the effectiveness of our CSVC framework in incorporating the assumed causal relationships.}
  \label{fig:figure5}
\end{figure*}

% === Figure 6 ===
\begin{figure}[ht]
  \centering
  \includegraphics[width=1.0\linewidth]{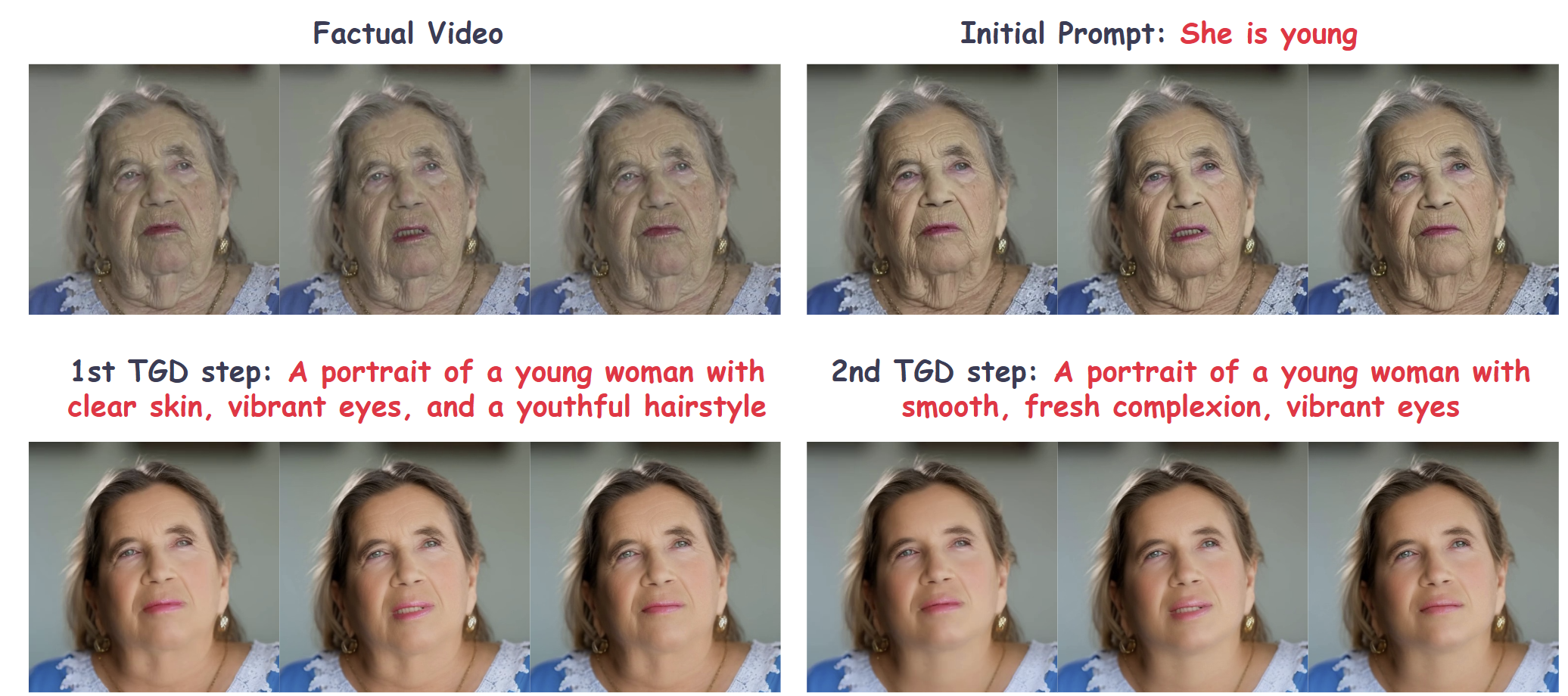}
  \caption{Counterfactual transformation of an elderly woman into a young woman (top row) through two iterative Textual Gradient Descent (TGD) steps~\cite{textgrad5optimizing} in the bottom row produced by our proposed CSVC with the FLATTEN~\cite{cong2024FLATTEN} editing method.}
  \label{fig:figure6}
\end{figure}

\subsubsection{Quantitative Evaluation.}

%We thoroughly assess the generated counterfactual videos utilizing metrics that capture key axiomatic properties of counterfactuals \cite{galles1998axiomatic, halpern2000axiomatizing}. From a causal perspective, we evaluate the produced video counterfactuals in terms of effectiveness \cite{monteiromeasuring, melistas2024benchmarking} and minimality \cite{melistas2024benchmarking, sanchez2022diffusion}. Moreover, we employ general video quality metrics including DOVER \cite{wu2023dover, liu2024evalcrafter} and FVD \cite{unterthiner2018towardsFVD} to assess visual fidelity and realism, as well as the CLIP \cite{radford2021learningCLIP} score between adjacent frames to validate semantic consistency and assess temporal coherence. We compare our method with both vanilla video editing methods, which use the initial counterfactual prompts from our evaluation dataset and a naive LLM-based paraphrasing baseline, in which an LLM takes the initial target counterfactual prompt as input and is instructed to rephrase it using more expressive language without having access to the counterfactual video. In addition, we report results for our approach both with and without augmenting the VLM evaluation instruction using the causal decoupling prompt (VLM loss w/o causal dec, VLM loss w/ causal dec), as described in Section ~\ref{VLM-base_loss}.

We evaluate the generated counterfactual videos using metrics that capture key axiomatic properties of counterfactuals \cite{galles1998axiomatic, halpern2000axiomatizing}, focusing on effectiveness \cite{monteiromeasuring, melistas2024benchmarking} and minimality \cite{melistas2024benchmarking, sanchez2022diffusion}. To assess visual fidelity and temporal coherence, we employ DOVER \cite{wu2023dover, liu2024evalcrafter}, FVD \cite{unterthiner2018towardsFVD}, and CLIP \cite{radford2021learningCLIP} score between adjacent frames. We compare CSVC against vanilla video editing baselines using the initial counterfactual prompts, an LLM-based paraphrasing baseline where an LLM rephrases the target counterfactual prompt, and report results with and without the causal decoupling prompt.
%\footnote{See the Methodology section and Appendix.} 
%(CSVC w/o causal dec, CSVC loss w/ causal dec.
%\subsubsection{Comparison of vanilla editing systems.}
%From Table \ref{all_metrics}, and specifically looking at the initial prompt rows, we observe that TokenFlow achieves the best balance between causal effectiveness and minimality compared to the other two baseline methods. Tune-A-Video generates reasonably effective counterfactuals but struggles with minimality, showing the worst values across all three frameworks in both LPIPS and VLM-based minimality metrics. Regarding overall video quality and temporal consistency, TokenFlow and FLATTEN outperform Tune-A-Video, indicating superior performance in maintaining visual coherence.

From Table \ref{all_metrics}, observing the initial prompt rows, TokenFlow achieves the best trade-off between causal effectiveness and minimality among the baselines. Tune-A-Video generates effective counterfactuals but performs worst in terms of minimality across both LPIPS and the VLM-based metric. In terms of overall video quality and temporal consistency, TokenFlow and FLATTEN outperform Tune-A-Video, maintaining stronger visual coherence.

\subsubsection{Effectiveness.}
To measure counterfactual effectiveness, we use VLMs prompted with multiple-choice questions on the intervened variables (age, gender, beard, bald). Table \ref{all_metrics} reports VLM accuracy for each variable under these interventions. CSVC improves causal effectiveness across all baseline methods, with the highest scores achieved when incorporating the causal decoupling prompt (CSVC loss w/ causal decoupling), indicating better steering toward counterfactuals that break strong causal relations (e.g., adding a beard to a female). While naive LLM paraphrasing occasionally boosts gender interventions for FLATTEN and TokenFlow, it generally fails due to hallucinations or irrelevant content that the diffusion model cannot handle.

\subsubsection{Minimality.}
%To measure the minimality of the interventions we utilize LPIPS \cite{zhang2018unreasonableLPIPS} and the VLM-based metric as described in \ref{para_3.5_minimality_VLM}. 
% Our method reveals the interplay between preserving proximity (as described in \ref{para_3.5_minimality_VLM}) to the factual video and adhering to the counterfactual text conditioning. From Table \ref{all_metrics}, we observe that the LPIPS metric tends to increase as the counterfactual edits become more effective. In addition, when measuring minimality using the proposed VLM-based metric, we observe a similar trend, as the cosine similarity of the transformer semantic embeddings is decreased slightly. Nevertheless, the deviations from baseline frameworks remain marginal. Overall, we can conclude that our approach is capable of achieving minimality scores comparable to those of baseline methods, thereby ensuring a reasonable balance between effectiveness and minimality.
To evaluate minimality, we use LPIPS \cite{zhang2018unreasonableLPIPS} and our proposed VLM-based metric (Equation \ref{eq:minimality}). Our results reveal the trade-off between preserving proximity to the factual video and adhering to the counterfactual text conditioning. As shown in Table \ref{all_metrics}, LPIPS increases as counterfactual edits become more effective, with the VLM-based metric showing a similar trend through slight decreases in embedding cosine similarity. However, deviations from baseline methods remain marginal, indicating that CSVC achieves minimality scores comparable to vanilla frameworks while maintaining a balance with causal effectiveness.

\subsubsection{Video Quality and Temporal Consistency.}
%Table \ref{all_metrics} presents quantitative results for general video quality (DOVER, FVD) and temporal consistency (CLIP \cite{radford2021learningCLIP} score). We observe that DOVER score \cite{wu2023dover}, which assesses generated counterfactual videos from both technical and aesthetic perspectives, shows only marginal differences between the baseline methods and our VLM-steering approach. In addition, the FVD \cite{unterthiner2018towardsFVD} score shows a slight increase for our proposed method, indicating that as the counterfactuals become more effective, they deviate more noticeably from the observational distribution. Lastly, the deviations in CLIP scores for temporal consistency are minimal compared to the vanilla methods. Overall, we conclude that the measurements of general video quality and temporal consistency metrics indicate no significant deviation from the baselines, demonstrating that our method improves counterfactual effectiveness without compromising other critical factors such as video realism and temporal coherence.
Table \ref{all_metrics} reports quantitative results for video quality (DOVER, FVD) and temporal consistency (CLIP \cite{radford2021learningCLIP}). DOVER \cite{wu2023dover} shows only minor differences between baselines and our CSVC framework. FVD \cite{unterthiner2018towardsFVD} increases slightly, reflecting greater deviation from the observational distribution as counterfactuals become more effective. CLIP-based temporal consistency remains close to the vanilla methods. Overall, our CSVC approach improves counterfactual effectiveness without compromising video realism or temporal coherence.

%Figures \ref{fig:figure4}
\subsubsection{Qualitative Evaluation}
%In Figure \ref{fig:figure5} we present qualitative results of our method across all three video editing systems: FLATTEN~\cite{cong2024FLATTEN}, Tune-A-Video~\cite{wu2023tune}, and TokenFlow~\cite{geyer2024tokenflow}. The first row depicts the factual video along with the factual (source) prompt, while the remaining rows show counterfactuals produced using the initial counterfactual prompt, the LLM-paraphrased prompt, and our causally optimized prompt. We observe that our approach effectively generates counterfactual videos that faithfully incorporate the desired interventions. Specifically, we can derive a broad range of counterfactuals -- from edits that break strong causal relationships (e.g., adding a beard to a woman) to age transformations (e.g., making an older man appear as a child) and gender transformations (e.g., making a man appear as a woman in Figure \ref{fig:figure5}). Furthermore, the results highlight the superiority of our proposed VLM causal steering compared to the naive prompt paraphrasing by an LLM.
Figure \ref{fig:figure5} shows \emph{qualitative results} \footnote{Due to space constraints, additional qualitative results are provided in the Appendix and supplementary materials.} across FLATTEN~\cite{cong2024FLATTEN}, Tune-A-Video~\cite{wu2023tune}, and TokenFlow~\cite{geyer2024tokenflow}. The top row displays the factual video and prompt, while subsequent rows show counterfactuals generated with the initial counterfactual prompt, an LLM-paraphrased prompt, and our causally optimized prompt with CSVC. Our framework produces counterfactuals that accurately reflect the desired interventions, including breaking strong causal relationships (e.g., adding a beard to a woman), as well as causally faithful age and gender transformations. The results also showcase the effectiveness of CSVC over naive LLM prompt paraphrasing. 
Figure \ref{fig:figure6} illustrates CSVC with the FLATTEN method, where iterative gradient steps (2nd row) guide generation toward the intended intervention (youthful appearance), demonstrating controllable causal steering.

\section{Conclusion}
\label{Discussion}
In this paper, we propose a causal framework, namely, CSVC, for steering text-guided one- and zero-shot LDM-based video editing systems to generate causally faithful video counterfactuals. Causal priors are encoded via target prompts that reflect relationships defined by a causal graph. Building on the insight that causal counterfactuals reside in the latent space of LDMs, CSVC leverages textual evaluative feedback from a VLM to iteratively refine the target causal prompt, guiding the LDM toward generating novel OOD counterfactuals. This optimization strategy offers a principled approach to counterfactual generation, enhancing causal alignment while preserving visual realism, minimality, and temporal coherence. 
Experimental results highlight the effectiveness and controllability of the proposed CSVC framework, underscoring its potential to advance causal reasoning in diffusion-based generative models.
Importantly, our findings demonstrate that diffusion models can be effectively steered to generate OOD counterfactuals.

\subsubsection{Acknowledgments.}
This work has been partially supported by project MIS 5154714 of the National Recovery and Resilience Plan Greece 2.0 funded by the European Union under the NextGenerationEU Program. S.A. Tsaftaris acknowledges support from the Royal Academy of Engineering and the Research Chairs and Senior Research Fellowships scheme (grant RCSRF1819/8/25), and the UK’s Engineering and Physical Sciences Research Council (EPSRC) support via grant EP/X017680/1, and the UKRI AI programme and EPSRC, for CHAI - EPSRC AI Hub for Causality in Healthcare AI with Real Data [grant number EP/Y028856/1]. Hardware resources were granted with the support of GRNET.

\bibliography{aaai2026}

\clearpage
%\begin{comment}
%\input{appendix}
%===========BEGIN APPENDIX=============================
\section*{Appendix}
%\section{Summary}
%The appendix includes a description of the dataset construction process (Appendix \ref{appendix:dataset}) and the detailed structure of the evaluation instruction used in the VLM-based loss (Appendix \ref{appendix:prompts}). We also outline the optimization procedure for improving prompt quality using textual feedback and textual gradients ($\frac{\partial \mathcal{L}}{\partial \mathcal{P}}$) provided by the VLM-based loss (Appendix \ref{appendix:VLM_loss_output}). Furthermore, we present the VLM evaluation pipelines used to assess effectiveness and minimality (Appendix~\ref{appendix:VLM_metrics}). Lastly, we include qualitative results (Appendix~\ref{appendix:qualitative}) that further demonstrate the capabilities of our proposed approach.

\section*{Evaluation Dataset}
\label{appendix:dataset}

\begin{figure}[ht]
  \centering
  \includegraphics[width=1.0\linewidth]{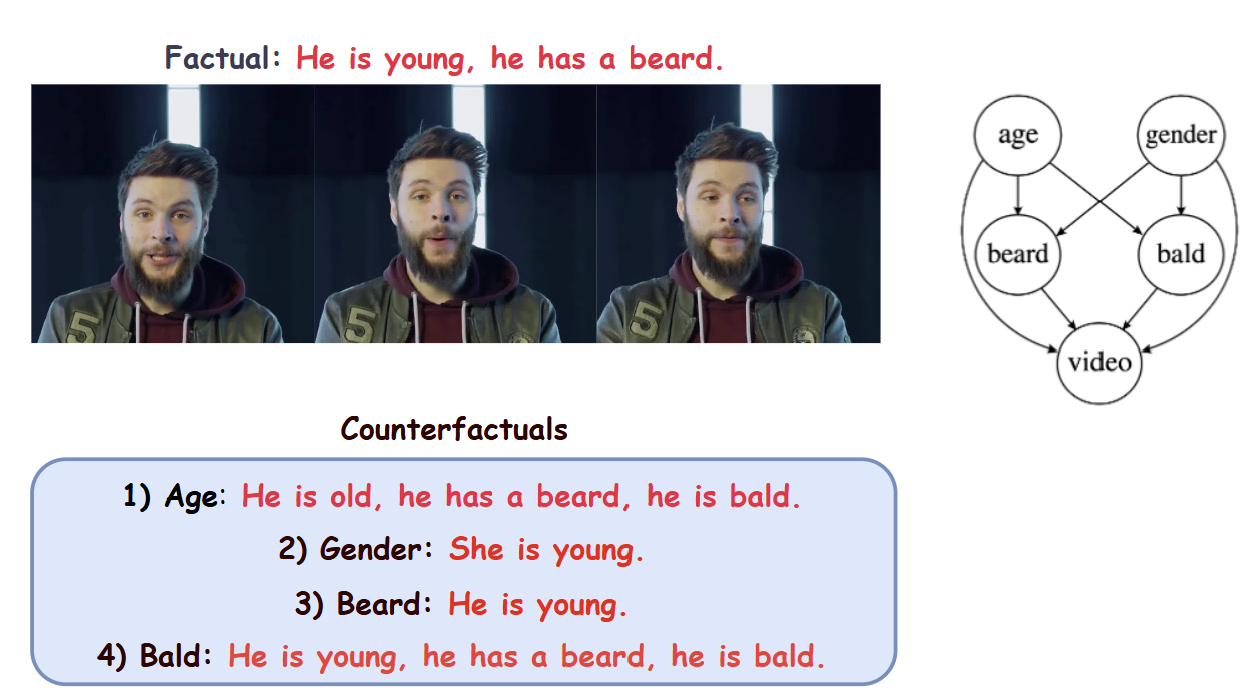}
  \caption{Evaluation dataset structure: Each factual prompt, sourced from CelebV-Text, is associated with four counterfactual prompts. Each counterfactual (target) represents an intervention on one of the following variables--age, gender, beard, or baldness. Interventions on upstream causal variables (e.g., age or gender) may lead to changes in downstream variables (e.g., beard or baldness), which are automatically incorporated into the counterfactual prompt.
}
  \label{figure7_updated}
\end{figure}

 We curated an evaluation dataset consisting of 67 text-video pairs sourced from the large-scale facial text–video dataset CelebV-Text~\cite{celebv-text}. We extracted the first 24 frames from each video and resized them to a resolution of 512$\times$512. Each video in CelebV-Text is associated with a text prompt describing static appearance attributes. We model the data-generating process using the causal graph shown in Figure~\ref{figure7_updated}. Given the factual (original) text prompt for each video, sourced from CelebV-Text~\cite{celebv-text}, we derive four counterfactual (target) prompts that are as similar as possible to the factual prompt, differing only in the specified interventions. To produce the counterfactual prompts and incorporate the interventions, we follow the assumed causal relationships depicted in the causal graph (Figure~\ref{figure7_updated})--for example, older men are more likely to have a beard or be bald than younger men, while women typically do not exhibit facial hair or baldness. Therefore, we successfully construct causal prompts. The counterfactual prompts are generated leveraging GPT-4 via the OpenAI API. An example is shown in Figure~\ref{figure7_updated}. 

\section*{Automatic Generation of Initial Counterfactual Causal Prompts}
In Listing \ref{lst:GPT_in_context}, we provide a part of the GPT-4 prompt used to derive the initial counterfactual prompts from the factual prompts for each video by incorporating the causal graph (Figure \ref{fig:figure_2}) and leveraging in-context learning \cite{dong2022survey_incontext}. To generate the 4 counterfactual prompts per video, we additionally supply GPT-4 with all 67 factual (source) descriptions of the original videos. In total, we produce 268 ($67\times4$) counterfactual prompts (four per video). The full prompt is included in our code. 

\begin{listing}[tb]%
\caption{GPT prompt for generating causally faithful counterfactual prompts}%
\label{lst:GPT_in_context}%
\begin{lstlisting}[basicstyle=\ttfamily\small, numbers=none]
You are given a causal DAG with 4 variables: age, gender, beard, and baldness.

Causal relationships:
- age -> beard
- age -> bald
- gender -> beard
- gender -> bald

Domain knowledge:
1. Older men are more likely to have a beard and be bald compared to younger men.
2. Men are more likely to have a beard and be bald compared to women.

Task:
Given a factual prompt that describes a person (e.g., He is young, he has a beard),
generate 4 counterfactual prompts by intervening on each variable (age, gender,beard, bald) while respecting the causal relationships.
Examples:
---
Factual:
He is young
Counterfactuals:
age: He is old, he has a beard, he is bald
gender: She is young
beard: He is young, he has a beard
bald: He is young, he is bald
---
Factual:
He is young, he has a beard
Counterfactuals:
age: He is old, he has a beard, he is bald
gender: She is young
beard: He is young
bald: He is young, he has a beard, he is bald
---
Factual:
He is old, he is bald
Counterfactuals:
age: He is young
gender: She is old
beard: He is old, he has a beard, he is bald
bald: He is old
---
Factual:
She is old
Counterfactuals:
age: She is young
gender: He is old, he has a beard, he is bald
beard: She is old, she has a beard
bald: She is old, she is bald

\end{lstlisting}
\end{listing}

\clearpage
\section*{Additional implementation details}
\label{appendix:dataset_implement}
For each baseline video editing method (FLATTEN~\cite{cong2024FLATTEN}, Tune-A-Video~\cite{wu2023tune}, and TokenFlow~\cite{geyer2024tokenflow}), we adopt the default experimental hyperparameters provided in the original works. In our experiments, we implement the VLM-based textual loss in our CSVC framework using the GPT-4o model via the OpenAI API. However, our approach is also compatible with local VLMs currently supported by the TextGrad package~\cite{textgrad5optimizing}. The LLM used to perform the TextGrad update (Equation \ref{TGD_update}) is GPT-4o--the same model used for the VLM loss. We also use the GPT-4o API to compute the VLM minimality metric, as it offers improved filtering of the causal graph variables in the generated text descriptions. In addition, for the BERT-based semantic text encoder $\tau_\phi$ used in Equation \ref{eq:minimality} to generate semantic text embeddings, we leverage the \textit{all-MiniLM-L6-v2} model~\cite{wang2020minilm}, which maps the text descriptions into a 384-dimensional vector space. Lastly, to evaluate effectiveness as expressed in Equation ~\ref{eq:effectiveness}, we utilize the \textit{llava-hf/llava-v1.6-mistral-7b-hf} model~\cite{li2024llava-next_}.

\section*{Prompts}
\label{appendix:prompts}
\subsection*{Evaluation Instruction}
We outline the methodology used to construct the evaluation instruction prompt for the VLM-based textual loss of the CSVC framework, as described in Section \ref{VLM-base_loss}. First, given the factual (source) prompt of the original video and the initial counterfactual (target) prompt--we programmatically extract the target interventions by comparing the two. In Listing \ref{lst:target_interventions}, we provide representative examples.

\begin{listing}[tb]%
\caption{Target Interventions Extraction}%
\label{lst:target_interventions}%
\begin{lstlisting}[basicstyle=\ttfamily\small, numbers=none]
Factual prompt: This woman is young.
Initial Counterfactual prompt: This woman is old.
Target interventions: old (age)

Factual prompt: He is young, he has a beard.
Initial Counterfactual prompt: She is young.
Target interventions: woman, no-beard (gender)

Factual prompt: This woman is young.
Initial Counterfactual prompt: This woman is young, she has a beard.
Target interventions: beard (beard)

Factual prompt: A man is young.
Initial Counterfactual prompt: A man is young, he is bald.
Target interventions: bald (bald)
\end{lstlisting}
\end{listing}

Given the initial counterfactual prompt and the target interventions, we provide the VLM with the following evaluation instruction:

\begin{listing}[tb]%
\caption{VLM Evaluation Instruction}%
\label{lst:VLM_evaluation_instruction}%
\begin{lstlisting}[basicstyle=\ttfamily\small, numbers=none]
- You are given an image of a person's face.

- A counterfactual target prompt is provided: {counterfactual_prompt}

- Corresponding interventions are specified: {target_interventions}

- Evaluate how well the given image aligns with the specified counterfactual attributes in the target prompt.

- Calculate an accuracy score based only on the attributes that were explicitly modified (i.e., the interventions).

- Do not describe or alter any other visual elements such as expression, hairstyle, background, clothing, or lighting.

- Identify and list any attributes from the interventions that are missing or incorrectly rendered.

- Criticize.

- Suggest improvements to the counterfactual prompt to better express the intended interventions.

- The optimized prompt should maintain a similar structure to the original prompt.

- If the alignment is sufficient, return: "No optimization is needed".
\end{lstlisting}
\end{listing}

%as described in Section~\ref{VLM-base_loss},
\subsection*{Causal decoupling prompt}
We further augment the evaluation instruction prompt with a causal decoupling prompt (Listing \ref{lst:decoupling_prompt}), 
in cases where interventions involve downstream variables (e.g., beard, bald) in the causal graph. This results in optimized prompts that exclude references to upstream variables (e.g., age, gender), effectively breaking the assumed causal relationships and simulating graph mutilation~\cite{pearl2009causality}. By using such prompts, the LDM backbone of the video editing method can generate OOD videos that violate the assumptions of the causal graph--for example, by adding a beard to a woman.

\begin{listing}[tb]%
\caption{Causal Decoupling Prompt}%
\label{lst:decoupling_prompt}%
\begin{lstlisting}[basicstyle=\ttfamily\small, numbers=none]
If either beard or bald appears in target_interventions, do not include references to age or gender.
\end{lstlisting}
\end{listing}

\clearpage
\section*{Evaluative Textual Feedback from VLM-Based Loss and Textual Gradient Computation}
For demonstration purposes, we provide the textual feedback from the VLM-based loss in our CSVC framework during prompt optimization for the first video in Figure ~\ref{figure1_updated} (transforming an old woman into a young one) with the TokenFlow \cite{geyer2024tokenflow} editing method. In addition, we present the corresponding textual gradient $\frac{\partial \mathcal{L}}{\partial \mathcal{P}}$, which is used to update the initial prompt via the TextGrad \cite{textgrad5optimizing}. First, we generate the counterfactual video using the initial counterfactual prompt (A woman is young), which represents an intervention on the age variable. Then, we provide a generated counterfactual frame to the VLM for evaluation.
\label{appendix:VLM_loss_output}

\begin{figure}[ht]
  \centering
  \includegraphics[width=1.0\linewidth]{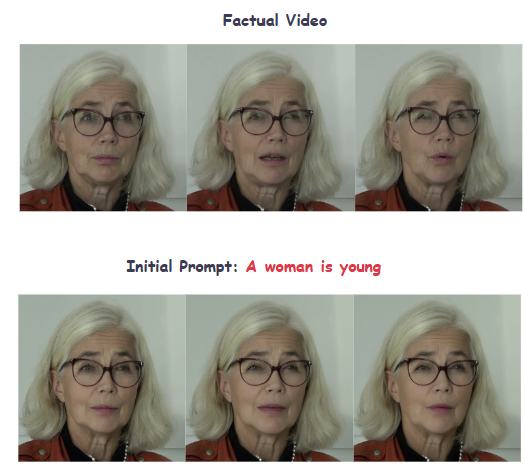}
  \caption{Counterfactual video generated using the initial prompt, which failed to incorporate the target intervention (young).}
  \label{figure8_updated}
\end{figure}

In Listing \ref{lst:VLM_feedback}, we present the textual feedback produced by the VLM given a frame from the ineffective generated counterfactual video of Figure \ref{figure8_updated}.

\begin{listing}[tb]%
\caption{VLM textual feedback from the first prompt optimization iteration}%
\label{lst:VLM_feedback}%
\begin{lstlisting}[basicstyle=\ttfamily\small, numbers=none]
The image does not align well with the specified counterfactual attributes from the target prompt. The intervention specified was "young," but the person depicted in the image appears to be an older adult, likely in her 60s or 70s, based on visible signs of aging such as wrinkles and grey hair.

**Accuracy Score:** 0/1 (The attribute "young" was not correctly rendered)

**Failed Attributes:**
- Youthfulness: The individual in the image does not appear young.

**Suggested Improvement to the Counterfactual Prompt:**
"A woman in her early twenties" 

This optimized prompt more specifically targets the desired age range, enhancing the likelihood that the generated image will meet the counterfactual criteria.
\end{lstlisting}
\end{listing}

In addition, in Listing \ref{lst:textual_gradient}, we display the textual gradients $\frac{\partial \mathcal{L}}{\partial \mathcal{P}}$, which contain detailed criticisms and suggestions for improving the initial prompt based on the VLM's textual feedback. To compute the textual gradient $\frac{\partial \mathcal{L}}{\partial \mathcal{P}}$, TextGrad leverages a backward engine (LLM/VLM) that is prompted with an augmented prompt derived by combining internal TextGrad system prompts, the evaluation instruction (Listing \ref{lst:VLM_evaluation_instruction}), the generated counterfactual frame, the optimized prompt $\mathcal{P}$, and the evaluative textual feedback (Listing \ref{lst:VLM_feedback}). In Listing \ref{lst:textgrad_prompt_for_grad} we present part of the internal prompt that TextGraD employs to produce textual gradients.

\begin{listing*}[tb]%
\caption{TextGrad Internal Prompt for Computing $\frac{\partial \mathcal{L}}{\partial \mathcal{P}}$}%
\label{lst:textgrad_prompt_for_grad}%
\begin{lstlisting}[basicstyle=\ttfamily\small, numbers=none]
You will give feedback to a variable with the following role: <ROLE> prompt to optimize </ROLE>. 
Here is an evaluation of the variable using a language model:

Above messages are the <LM_INPUT>

<LM_SYSTEM_PROMPT> 
You are an evaluation system that evaluates image-related questions. 
</LM_SYSTEM_PROMPT>

<LM_OUTPUT> 
The image does not align well with the specified counterfactual attributes from the target prompt. The intervention specified was "young," but the person depicted in the image appears to be an older adult, likely in her 60s or 70s, based on visible signs of aging such as wrinkles and grey hair.

**Accuracy Score:** 0/1 (The attribute "young" was not correctly rendered)

**Failed Attributes:**
- Youthfulness: The individual in the image does not appear young.

**Suggested Improvement to the Counterfactual Prompt:**
"A woman in her early twenties" 

This optimized prompt more specifically targets the desired age range, enhancing the likelihood that the generated image will meet the counterfactual criteria.
</LM_OUTPUT>

<OBJECTIVE_FUNCTION>
Your goal is to give feedback and criticism to the variable given the above evaluation output. 
Our only goal is to improve the above metric, and nothing else. 
</OBJECTIVE_FUNCTION>

We are interested in giving feedback to the prompt to optimize for this conversation. 
Specifically, give feedback to the following span of text:

<VARIABLE> A woman is young </VARIABLE>

Given the above history, describe how the prompt to optimize could be improved to 
improve the <OBJECTIVE_FUNCTION>. Be very creative, critical, and intelligent.
\end{lstlisting}
\end{listing*}

\begin{listing*}[tb]%
\caption{Textual gradient $\frac{\partial \mathcal{L}}{\partial \mathcal{P}}$}%
\label{lst:textual_gradient}%
\begin{lstlisting}[basicstyle=\ttfamily\small, numbers=none]
The variable "A woman is young" lacks specificity and clarity, which likely contributed to the misalignment between the generated image and the intended attribute of youthfulness. The term "young" is highly subjective and can be interpreted variably depending on cultural, personal, and contextual factors. This ambiguity in the prompt may have led the Text-to-Image model to generate an image that does not meet the counterfactual criteria specified.
 
 **Feedback and Criticism:**
 
 1. **Specificity in Age Description**: The term "young" should be replaced with a more precise age or age range to guide the model more effectively. For instance, specifying "a woman in her early twenties" or "a woman aged 20-25 years" would provide a clearer target for the model, reducing ambiguity and increasing the likelihood of generating an image that aligns with the youthful attribute desired.
 
 2. **Inclusion of Contextual Details**: While the prompt should remain focused on the age attribute, adding minimal contextual details that align with youthful appearances could enhance the effectiveness of the prompt. For example, mentioning "a youthful demeanor" or "fresh-faced" could cue the model to focus on features typically associated with younger individuals.
 
 3. **Avoiding Overly Broad Descriptors**: The use of broad descriptors like "young" without additional qualifiers can lead to a wide range of interpretations. It's beneficial to avoid these where precision is required for generating specific outcomes.
 
 4. **Testing and Iteration**: After adjusting the prompt, it would be advisable to test the new prompt with the model to see if the changes yield a more accurate representation. This iterative approach can help fine-tune the prompt based on actual outcomes.
 
 By addressing these points, the prompt "A woman is young" can be optimized to more effectively communicate the desired attribute of youthfulness to the Text-to-Image model, thereby improving the alignment of the generated image with the counterfactual target.
\end{lstlisting}
\end{listing*}

The textual gradients $\frac{\partial \mathcal{L}}{\partial \mathcal{P}}$ (Listing \ref{lst:textual_gradient}) are provided as input to Textual Gradient Descent~\cite{textgrad5optimizing}, which leverages an LLM to update the optimized variable (prompt), as described in Equation \ref{TGD_update}. For simplicity and robustness in our experiments, we use the same LLM/VLM (GPT-4) for all operations: producing textual evaluative feedback, computing textual gradients, and updating the prompt with Textual Gradient Descent. After the TGD update the counterfactual prompt becomes: \textit{A woman in her early 20s with vibrant expression}.

\begin{figure}[ht]
  \centering
  \includegraphics[width=1.0\linewidth]{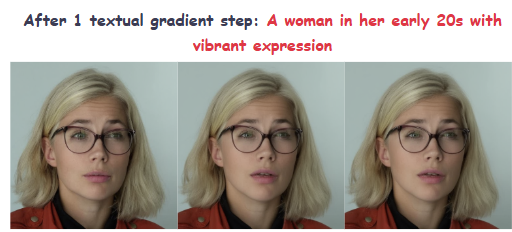}
  \caption{Counterfactual video generated using the optimized prompt, which successfully incorporates the target intervention (young).}
  \label{figure9_updated}
\end{figure}

\begin{listing*}[tb]%
\caption{VLM feedback from the second prompt optimization iteration}%
\label{lst:VLM_feedback_2nd_iter}%
\begin{lstlisting}[basicstyle=\ttfamily\small, numbers=none]
The input frame aligns well with the specified counterfactual attribute of appearing "young." The individual in the image presents as a young adult, which matches the intervention target of portraying youth. Therefore, the accuracy score based on the attribute of appearing young is high.

No attributes from the interventions failed to appear or were incorrectly rendered in this context.

Since the image successfully aligns with the desired attribute of youth, there is no need for optimization of the prompt. The response is "no_optimization".
\end{lstlisting}
\end{listing*}

In Listing~\ref{lst:VLM_feedback_2nd_iter}, we display the textual feedback from the VLM after providing it with a frame from the effective counterfactual video generated using the optimized prompt (Figure~\ref{figure9_updated}). With this prompt, the age intervention (young) is successfully incorporated. Consequently, the VLM returns a "no optimization" response, and the prompt optimization process terminates.

\clearpage
\clearpage
\section*{VLM-based metrics for Assessing Effectiveness and Minimality}
\label{appendix:VLM_metrics}
\subsection*{Effectiveness}
We present the VLM pipeline for evaluating causal effectiveness. As shown in Figure \ref{figure10_updated}, the VLM receives as input the generated counterfactual frame and a multiple-choice question--extracted from the counterfactual prompt that corresponds to the intervened attribute. Since we edit static attributes, a single frame is sufficient to assess the effectiveness of the interventions. An accuracy score is calculated across all generated counterfactual frames for each intervened variable (age, gender, beard, baldness) (Equation \ref{eq:effectiveness}).

\begin{figure}[ht]
  \centering
  \includegraphics[width=1.0\linewidth]{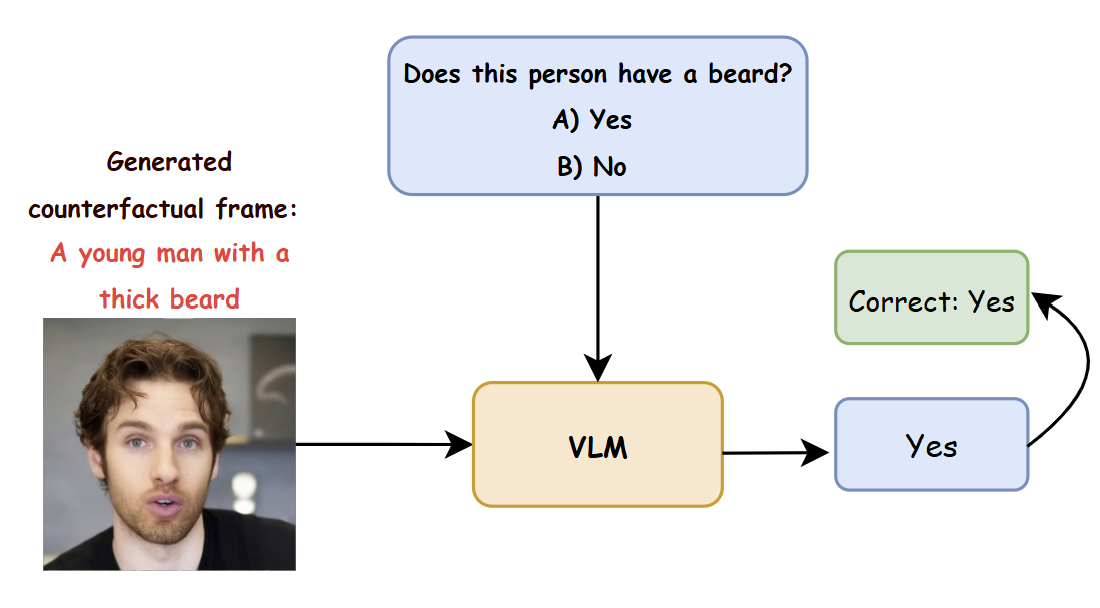}
  \caption{VLM causal effectiveness pipeline: example of a beard intervention.}
  \label{figure10_updated}
\end{figure}

%\vspace{2cm}	
\subsection*{Minimality}
\begin{figure}[ht]
  \centering
  \includegraphics[width=1.0\linewidth]{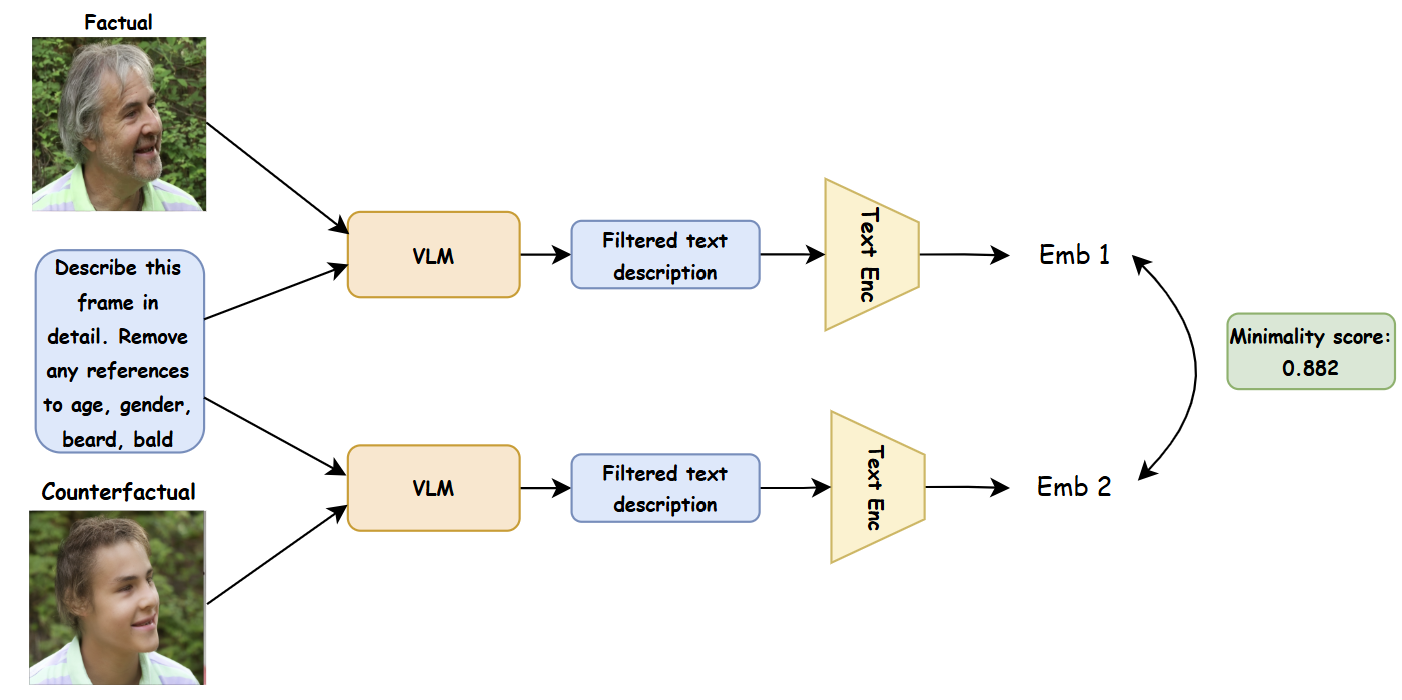}
  \caption{VLM minimality pipeline: example of a gender intervention.}
  \label{figure11_updated}
\end{figure}

In Figure \ref{figure11_updated}, we showcase the VLM pipeline for evaluating minimality (Equation \ref{eq:minimality}). The VLM takes as input frames extracted from the factual and counterfactual videos and produces text descriptions that exclude attributes from the causal graph. These text descriptions are then passed through a BERT-based semantic encoder~\cite{wang2020minilm} to generate semantic embeddings. The final minimality score is computed as the cosine similarity between these embeddings.
The exact prompt used to instruct the VLM to filter the text descriptions from the causal graph variables is provided in Listing \ref{lst:VLM_min_prompt}.

In Figure ~\ref{figure10_updated}, we display the filtered text descriptions produced by the VLM. This specific factual and counterfactual pair achieves a VLM minimality score of $0.882$. We observe that by measuring the semantic similarity of the VLM-generated text descriptions, we can isolate factors of variation not captured by the causal graph and effectively measure their changes under interventions on the causal graph variables. 
\begin{figure}[ht]
  \centering
  \includegraphics[width=1.0\linewidth]{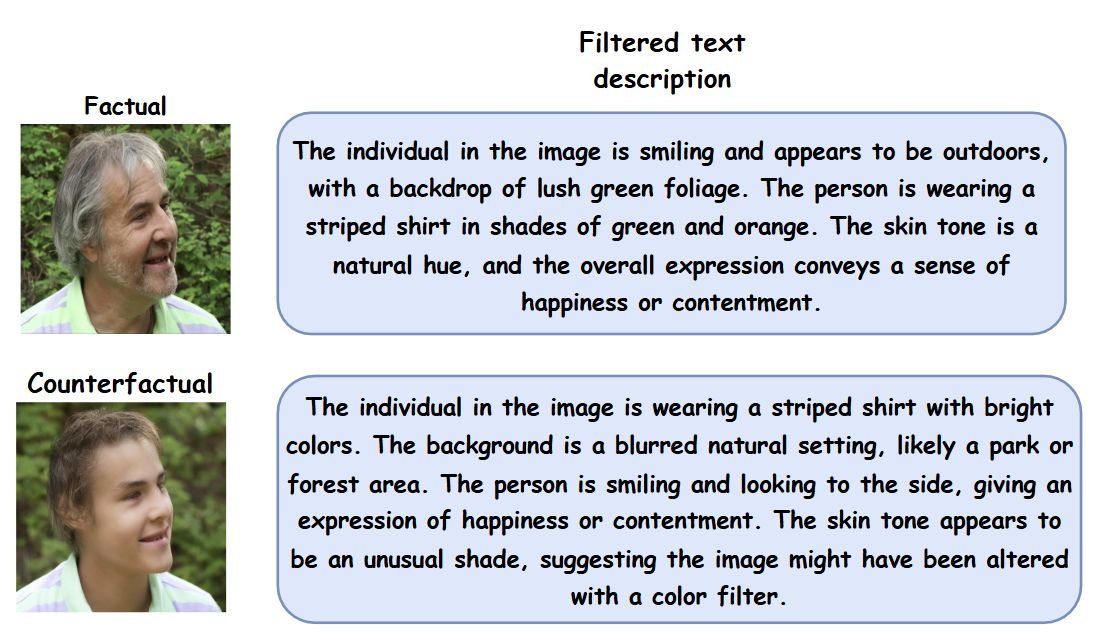}
  \caption{Filtered text descriptions derived from the VLM}
  \label{figure10_updated}
\end{figure}

\begin{listing}[tb]%
\caption{VLM Minimality Prompt}%
\label{lst:VLM_min_prompt}%
\begin{lstlisting}[basicstyle=\ttfamily\small, numbers=none]
Remove any references to age, gender (man, woman, he, she), beard, hair (including hairstyle, color, style, and facial hair), and baldness from the description.

Return only the filtered version of the text, without commentary or formatting.
.
\end{lstlisting}
\end{listing}

%\clearpage

\begin{figure*}[ht]
  \centering
  \includegraphics[width=1.0\linewidth]{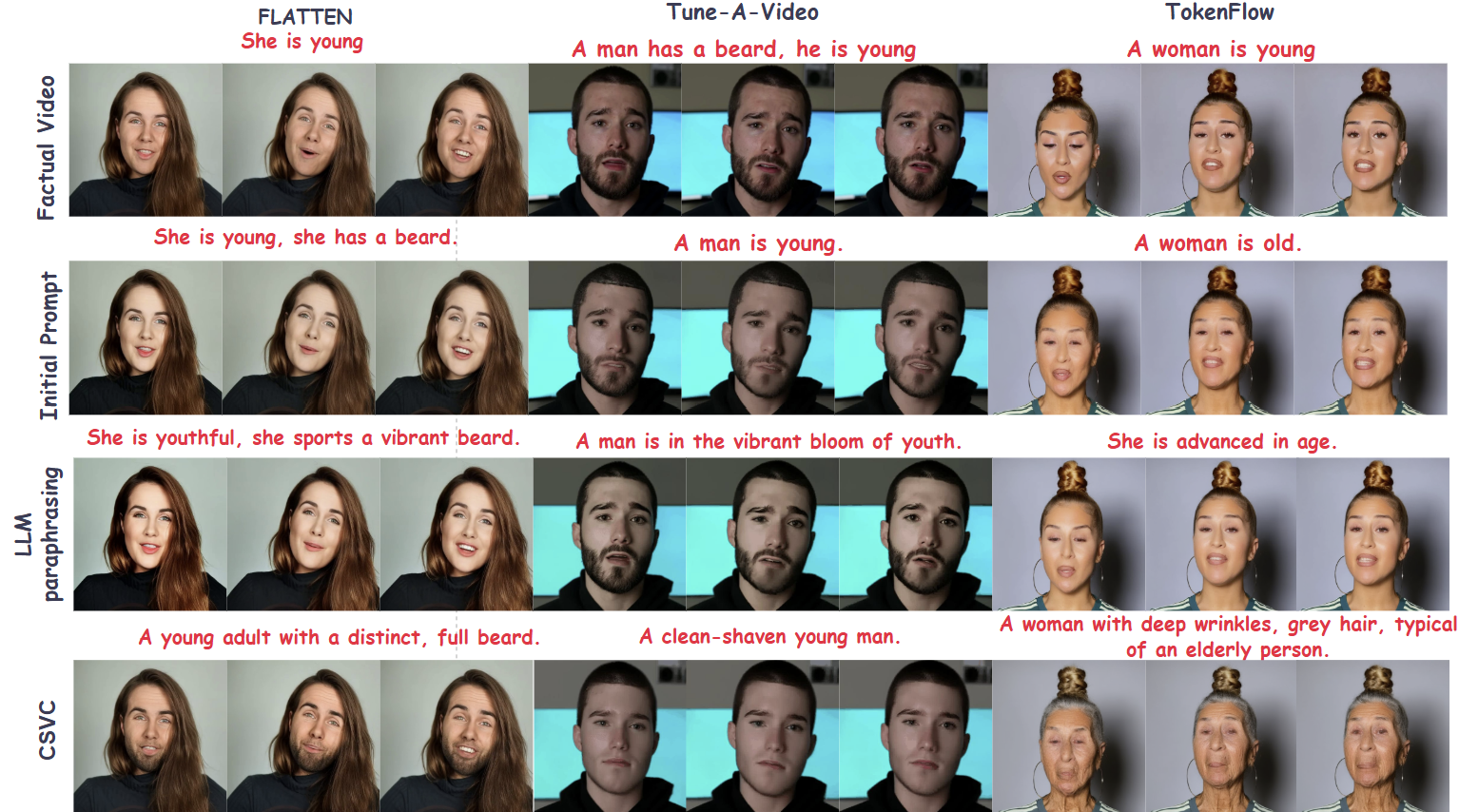}
 \caption{\textbf{Qualitative results}: Generated counterfactual videos illustrate the positive effect of our proposed CSVC framework (bottom row) when applied to recent video editing systems (FLATTEN~\cite{cong2024FLATTEN}, Tune-A-Video~\cite{wu2023tune}, and TokenFlow~\cite{geyer2024tokenflow}). \textbf{First panel:} intervention on beard (adding a beard to a woman). \textbf{Second panel:} intervention on beard (removing a beard from a man). \textbf{Third panel:} intervention on age (aging a woman).}
  \label{fig:figure4}
\end{figure*}

\begin{figure*}[ht]
  \centering
  \includegraphics[width=0.9\linewidth]{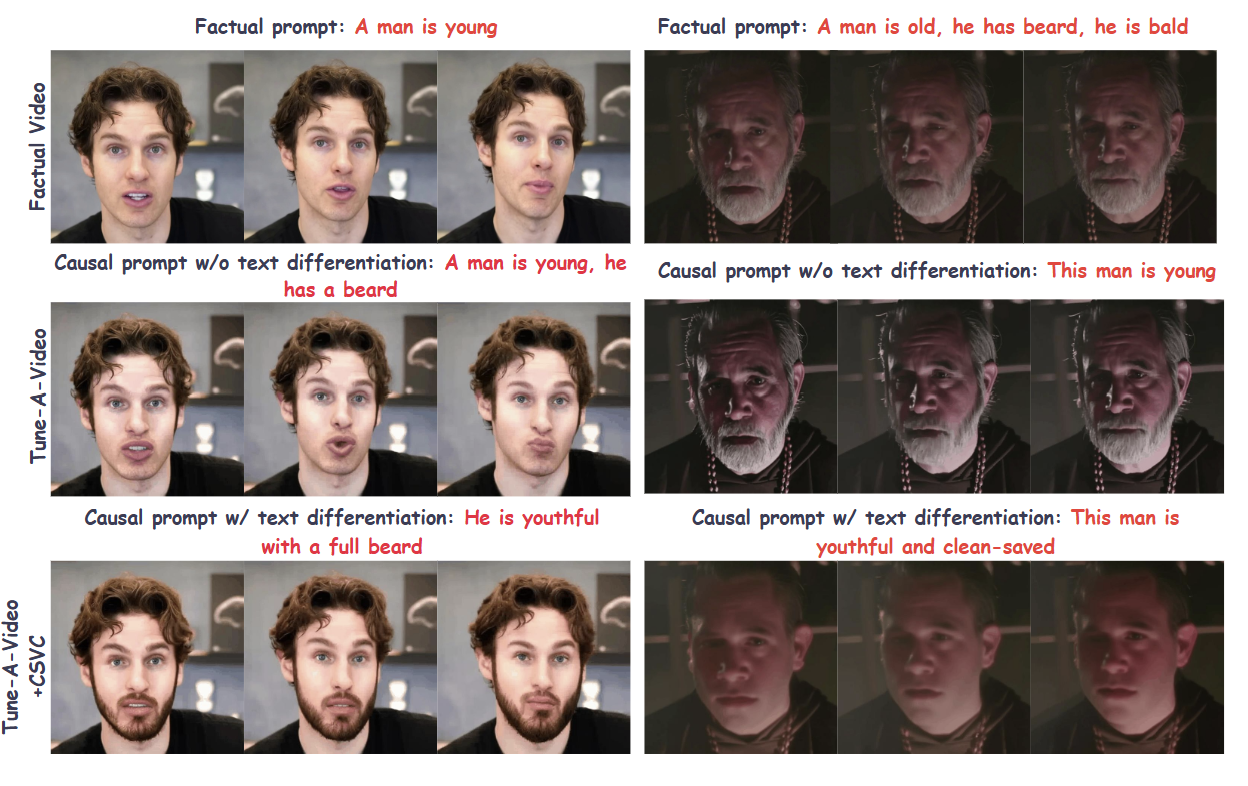}
  \caption{First panel: intervention on beard. Second panel: intervention on age.}
  \label{figure12_updated}
\end{figure*}

\label{appendix:qualitative}

\begin{figure*}[ht]
  \centering
  \includegraphics[width=0.9\linewidth]{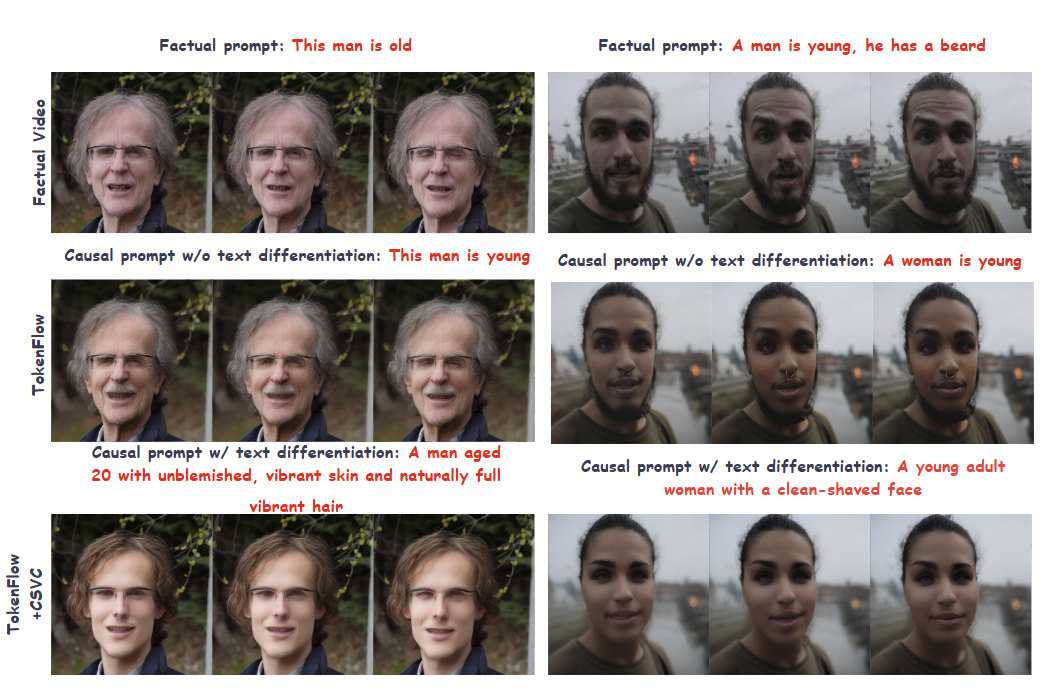}
  \caption{First panel: intervention on age. Second panel: intervention on gender.}
  \label{figure13_updated}
\end{figure*}

\begin{figure*}[ht]
  \centering
  \includegraphics[width=1.0\linewidth]{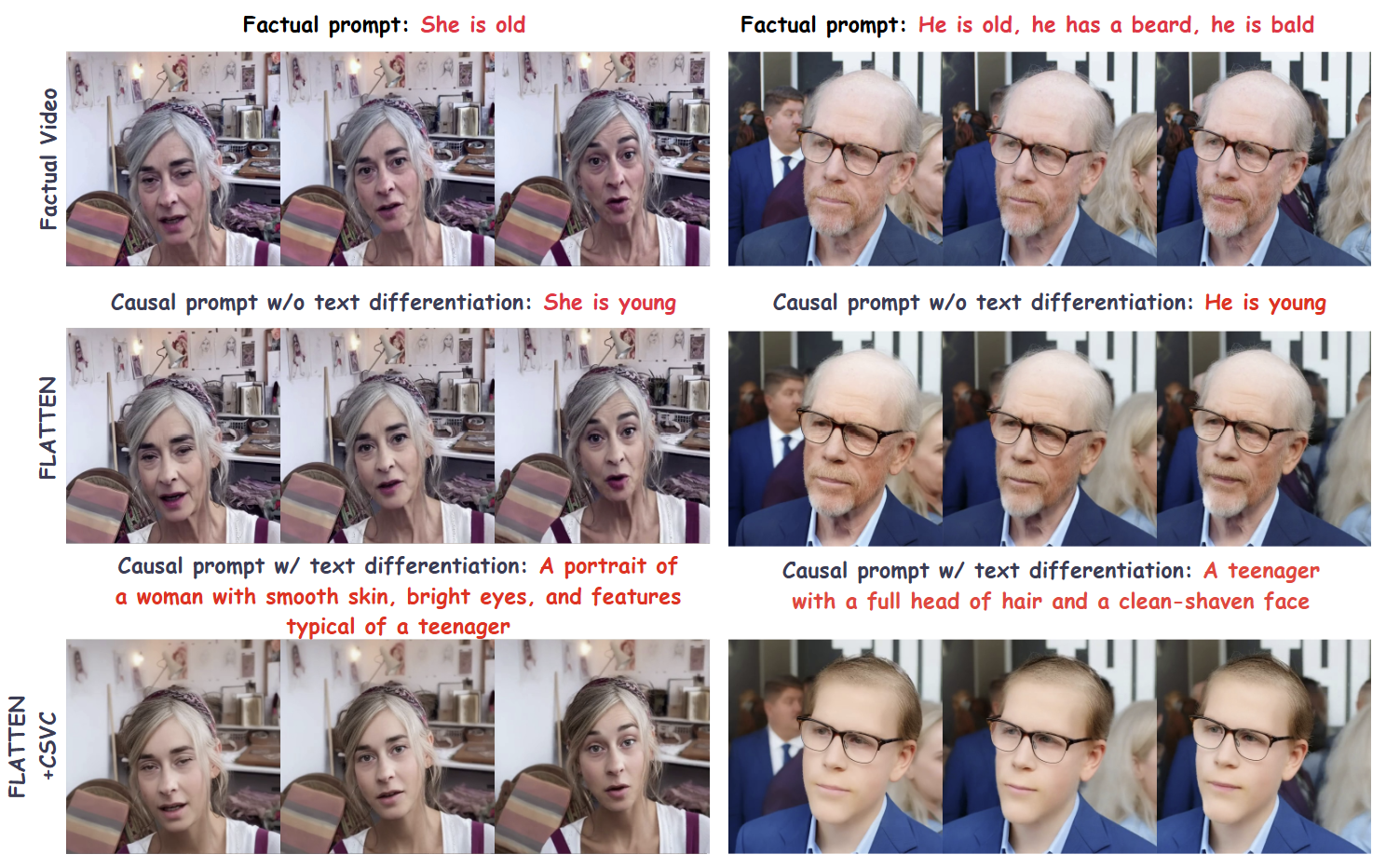}
  \caption{Interventions on age.}
  \label{figure14_updated}
\end{figure*}

%\newpage
%\vspace{1cm}	

%\subsection{More qualitative results with VLM Causal Steering}

\begin{figure*}[h]
  \centering
  \includegraphics[width=1.0\linewidth]{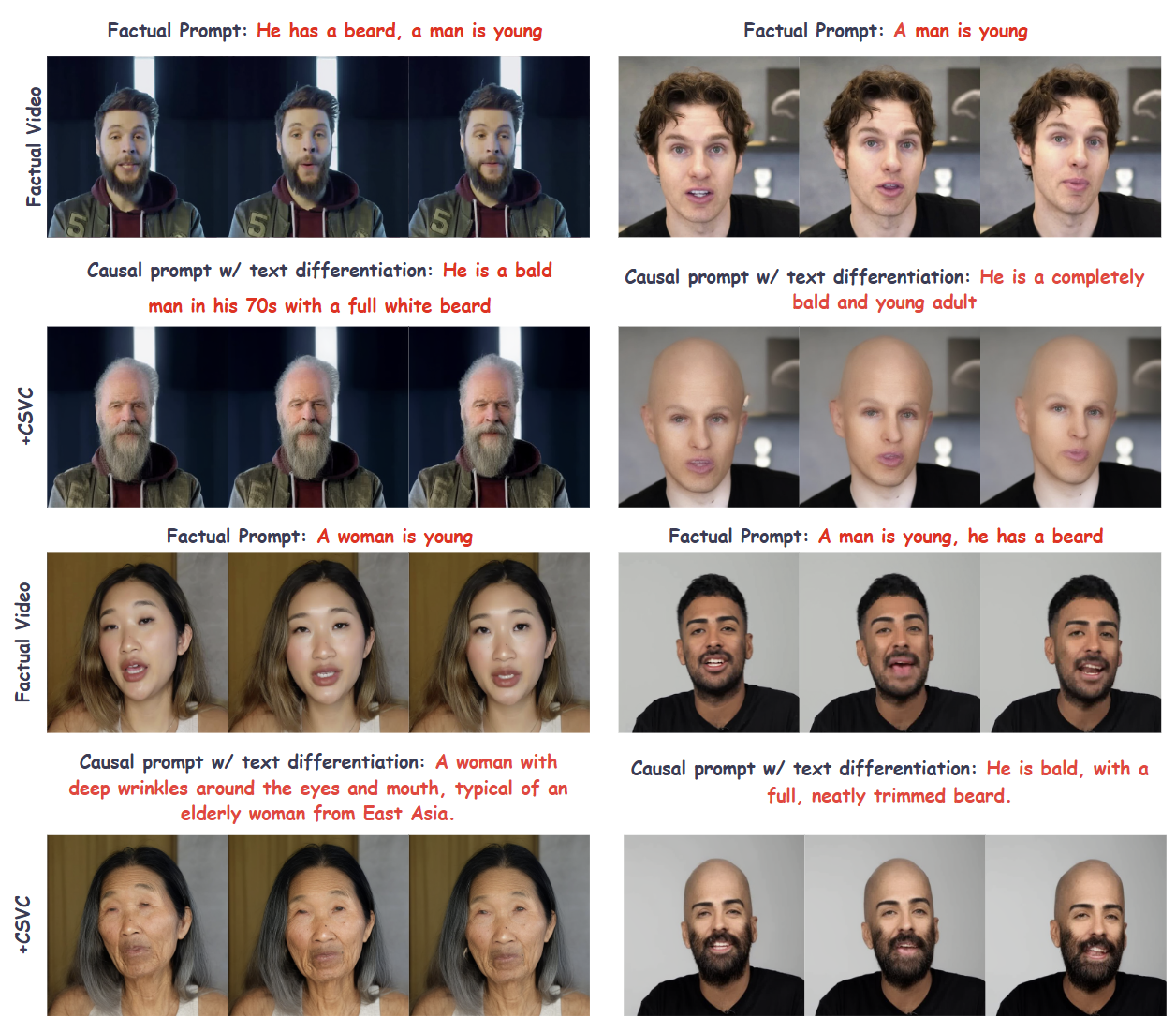}
  \caption{First panel: Interventions on age. Second panel: Interventions on baldness}
  \label{figure15_updated}
\end{figure*}

\section*{More qualitative results}
\label{appendix:qualitative}

In Figures \ref{fig:figure4}, \ref{figure12_updated}, \ref{figure13_updated}, \ref{figure14_updated}, and \ref{figure15_updated}, we present additional qualitative results generated using our proposed framework, "\textbf{C}ausal \textbf{S}teering for \textbf{V}ideo \textbf{C}ounterfactuals" (CSVC), with diffusion-based video editing systems for counterfactual generation.

\clearpage
\section*{Limitations}
We do not particularly add any loss to enforce temporal consistency beyond what each baseline method does. It is quite possible that static interventions on the attributes could alter temporal consistency but we haven't observed it in our case. In video editing, the ability to manipulate temporal attributes such as actions or dynamic scenes is crucial. Constructing such graphs and datasets are necessary to develop and test such methods and are left for future work.

\section*{Broader Impact}
Our framework (CSVC) for generating causally faithful video counterfactuals enhances video synthesis, interpretable AI, and content manipulation by providing better controllable edits. This could improve automated content generation in fields like healthcare (e.g., simulating treatment outcomes or disease progression under varied causal conditions), education (e.g., allowing students to observe video counterfactuals of complex processes, such as surgical procedures or engineering designs), and digital media (e.g., enabling creative content manipulation). Furthermore, it can potentially address ethical concerns, regarding thoroughly evaluating the misuse of deepfake technologies, highlighting the need for responsible guidelines and safeguards.
%\end{comment}
\clearpage
%\input{ReproducibilityChecklist/LaTeX/ReproducibilityChecklist}

\end{document}